\documentclass{article}

\usepackage{microtype}
\usepackage{graphicx}
\usepackage{subcaption}
\usepackage{booktabs} 
\usepackage{caption}
\usepackage{multirow}
\usepackage{pifont}
\usepackage{tcolorbox}
\usepackage{xcolor}
\tcbuselibrary{skins}
\usepackage{algorithm}
\usepackage{algorithmic}
\usepackage{ulem}

\usepackage{hyperref}

\usepackage[preprint]{icml2026}

\usepackage{amsmath}
\usepackage{amssymb}
\usepackage{mathtools}
\usepackage{amsthm}
\usepackage{xspace}
\usepackage{bbm}

\usepackage[capitalize,noabbrev]{cleveref}
\usepackage[toc,page,header]{appendix}

\input{extra_command/cmd_math.tex}

\theoremstyle{plain}

\theoremstyle{definition}

\theoremstyle{remark}

\newcommand{\ours}{DoTS\xspace}

\usepackage[textsize=tiny]{todonotes}

\icmltitlerunning{Decoupled Test-time Synthesis of SFT and RLVR Task Vectors}

\begin{document}

\twocolumn[
  \icmltitle{Decouple before Integration: Test-time Synthesis of SFT and RLVR Task Vectors}




  \begin{icmlauthorlist}
    \icmlauthor{Chaohao Yuan}{cuhk,damo}
    \icmlauthor{Chenghao Xiao}{damo}
    \icmlauthor{Yu Rong}{damo,hupan}
    \icmlauthor{Hong Cheng}{cuhk}
    \icmlauthor{Long-Kai Huang$^\dagger$}{hkbu}
  \end{icmlauthorlist}

  \icmlaffiliation{cuhk}{Department of Systems Engineering and Engineering Management, Chinese University of Hong Kong, Hong Kong, China}
  \icmlaffiliation{hkbu}{Department of Computer Science, Hong Kong Baptist University, Hong Kong, China}
  \icmlaffiliation{damo}{DAMO Academy, Alibaba Group, Hangzhou, China}
  \icmlaffiliation{hupan}{Hupan Lab, Hangzhou, China}

  \icmlcorrespondingauthor{Long-Kai Huang}{longkai@comp.hkbu.edu.hk}

  \icmlkeywords{Machine Learning, ICML}

  \vskip 0.3in
]



\printAffiliationsAndNotice{}  

\begin{abstract}
SFT and RLVR represent two fundamental yet distinct paradigms for LLM post-training, each excelling in distinct dimensions. SFT expands knowledge breadth while RLVR enhances reasoning depth. Yet integrating these complementary strengths remains a formidable challenge. Sequential training can cause catastrophic forgetting, and joint optimization often suffers from severe gradient conflicts.
We analyze SFT and RLVR through the lens of task vectors and reveal three structural properties behind these failures: a $\sim$30$\times$ magnitude disparity, $\sim$45\% sign interference, and heterogeneous module-wise update distributions. These findings show SFT and RLVR are difficult to integrate directly, but they also suggest that the two paradigms modify partly complementary components of the model. Motivated by these observations, we propose \textbf{D}ec\textbf{o}upled \textbf{T}est-time \textbf{S}ynthesis (\ours), a post-hoc framework allows SFT and RLVR checkpoints to be trained independently and synthesizes their capabilities only at inference time via task vector arithmetic, without updating model parameters.
To reduce interference, \ours applies selective sparsification with norm-preserving rescaling. It then uses Bayesian optimization on a small set of unlabeled queries to search for combination coefficients on the Pareto frontier of consistency and perplexity.
Empirically, \ours matches or exceeds the performance of training-based SFT--RLVR integration methods across multiple mathematical reasoning benchmarks, incurring only $\sim$3\% of the computational cost. When applied to stronger post-trained checkpoints, \ours surpasses SOTA models and generalizes to out-of-domain benchmarks without re-tuning. Code is available at \href{https://github.com/chaohaoyuan/DoTS}{chaohaoyuan/DoTS}.
\end{abstract}

\section{Introduction}

Supervised Fine-Tuning (SFT)~\cite{ouyang2022training, zhang2024when} and Reinforcement Learning with Verifiable Rewards (RLVR)~\cite{cobbe2021training, guo2025deepseek} represent two widely used paradigms for post-training Large Language Models (LLMs). They improve models in different but complementary ways. 
SFT expands knowledge breadth through direct supervision, enabling models to acquire task-specific and factual knowledge~\cite{gudibande2024the, allen-zhu2025physics}. RLVR amplifies reasoning depth via iterative refinement with verifiable feedback~\cite{lightman2024lets}.
Integrating these complementary strengths is essential for simultaneously improving the knowledge and reasoning capabilities in LLMs. 

Existing efforts to integrate SFT and RLVR capabilities mainly follow two routes. 
The first is sequential training~\cite{wen2025light, ma2025learning}, where a model is first trained with SFT to acquire domain knowledge and then further optimized with RLVR to improve reasoning. 
The second is unified training~\cite{yan2025learning}, where supervised learning signals are incorporated into the RLVR objective so that knowledge learning and reasoning optimization are performed together.

However, both routes are difficult to use reliably. 
In sequential training, the RLVR stage can overwrite or weaken the knowledge acquired during SFT, leading to catastrophic forgetting~\cite{rajani2025scalpel, chu2025sft}. 
In unified training, the supervised and reinforcement learning objectives may produce conflicting optimization signals, making it difficult to balance knowledge acquisition and reasoning improvement. 
Moreover, directly injecting supervised answers into RLVR may encourage the model to rely on memorized patterns instead of developing stronger reasoning behavior. 
These challenges raise an important question: \textit{are the failures of SFT--RLVR integration caused only by imperfect training recipes, or do they reflect a deeper structural incompatibility between the two paradigms?}

To answer this question, we analyze SFT and RLVR through the lens of \textit{task vectors}~\cite{ilharco2023editing}, defined as the parameter-space differences between post-trained and base models. 
Prior works have reported behavioral and geometric differences between SFT and RLVR~\cite{chu2025sft, rajani2025scalpel, zhu2025path, matsutani2025rl}. We complement these studies with a task-vector-level analysis and identify three structural properties that help explain why existing integration methods often fail. 
\textbf{(1) Extreme magnitude disparity.} The L2 norm of the SFT task vector is approximately $30\times$ larger than that of the RLVR task vector across layers. As a result, direct linear combination is dominated by SFT, while the weaker but important RLVR reasoning signal can be overwhelmed. 
\textbf{(2) Severe sign interference.} Without any processing, 44.91\% of parameters have opposite signs in the two task vectors. This means that nearly half of the parameters receive conflicting update directions, which helps explain the persistent gradient conflicts observed in joint training. 
\textbf{(3) Heterogeneous module-wise distributions.} SFT concentrates important updates in LayerNorm modules (19.9\%), whereas RLVR distributes its updates more broadly across attention, LayerNorm, and the LM head. This suggests that the two paradigms do not simply compete for the same parameters, but also affect partly complementary components of the model.

These findings suggest that the difficulty of SFT--RLVR integration is not only caused by imperfect training recipes. It also reflects a structural incompatibility between the two paradigms at the parameter level. At the same time, the heterogeneous module-wise distributions indicate that SFT and RLVR may encode complementary capabilities. 
If their interference can be controlled, combining them should be beneficial. 
This motivates a different integration strategy. Instead of forcing the two paradigms to coexist during training, we let each follow its own training objective independently and synthesize their capabilities at test time.

We propose \textbf{D}ec\textbf{o}upled \textbf{T}est-time \textbf{S}ynthesis (\textbf{\ours}), a framework that composes SFT and RLVR capabilities through task vector arithmetic at inference time.
To address the magnitude disparity and sign interference identified above, \ours applies \textit{selective sparsification}, which retains only the top-$k\%$ parameters by magnitude and rescales the remaining task vector to preserve its norm. This reduces sign interference from 44.91\% to as low as 7.1\%. 
We then use Bayesian optimization on a small set of unlabeled queries to search for the combination coefficients, guided by consistency and perplexity. Unlike training-based integration methods, \ours does not update model parameters. It only optimizes two scalar coefficients, $\lambda_{\text{SFT}}$ and $\lambda_{\text{RLVR}}$, requiring about 20 GPU hours compared with more than 600 GPU hours for training-based alternatives.

Empirically, \ours matches or exceeds training-based SFT--RLVR integration methods across multiple mathematical reasoning benchmarks, including AIME 2024/2025, AMC, MATH500, and OlympiadBench. It achieves this performance with only about 3\% of the computational cost. When applied to stronger post-trained checkpoints, ExGRPO~\cite{zhan2026exgrpo} and ReLIFT~\cite{ma2025learning}, \ours reaches an average score of 50.6, outperforming the strong LUFFY baseline by 1.4 points. Moreover, the coefficients learned on mathematical reasoning problems transfer directly to out-of-domain QA benchmarks, including ARC-C, GPQA, and MMLU-Pro, without re-tuning. This suggests that \ours captures a transferable mode of capability fusion rather than a task-specific heuristic.

Our main contributions are summarized as follows:
\begin{itemize}
\setlength{\itemsep}{0pt}
\vspace{-0.5em}

    \item We analyze the structural incompatibility between SFT and RLVR at the task-vector level. The analysis reveals a $\sim$30$\times$ magnitude disparity, $\sim$45\% sign interference, and heterogeneous module-wise update distributions, which help explain why naive merging and training-based integration can fail.

    \item We propose \ours, a decoupled test-time synthesis framework that combines independently trained SFT and RLVR checkpoints through task-vector arithmetic. The framework reduces interference through selective sparsification and selects combination coefficients through Bayesian optimization guided by consistency and perplexity.

    \item We show that \ours matches or exceeds training-based alternatives while reducing computational overhead by approximately 97\%. The learned synthesis coefficients also transfer to out-of-domain QA benchmarks without re-tuning.
\end{itemize}

\begin{figure*}[t]
    \centering
    \begin{subfigure}[t]{0.31\textwidth}
        \centering
        \includegraphics[width=\linewidth]{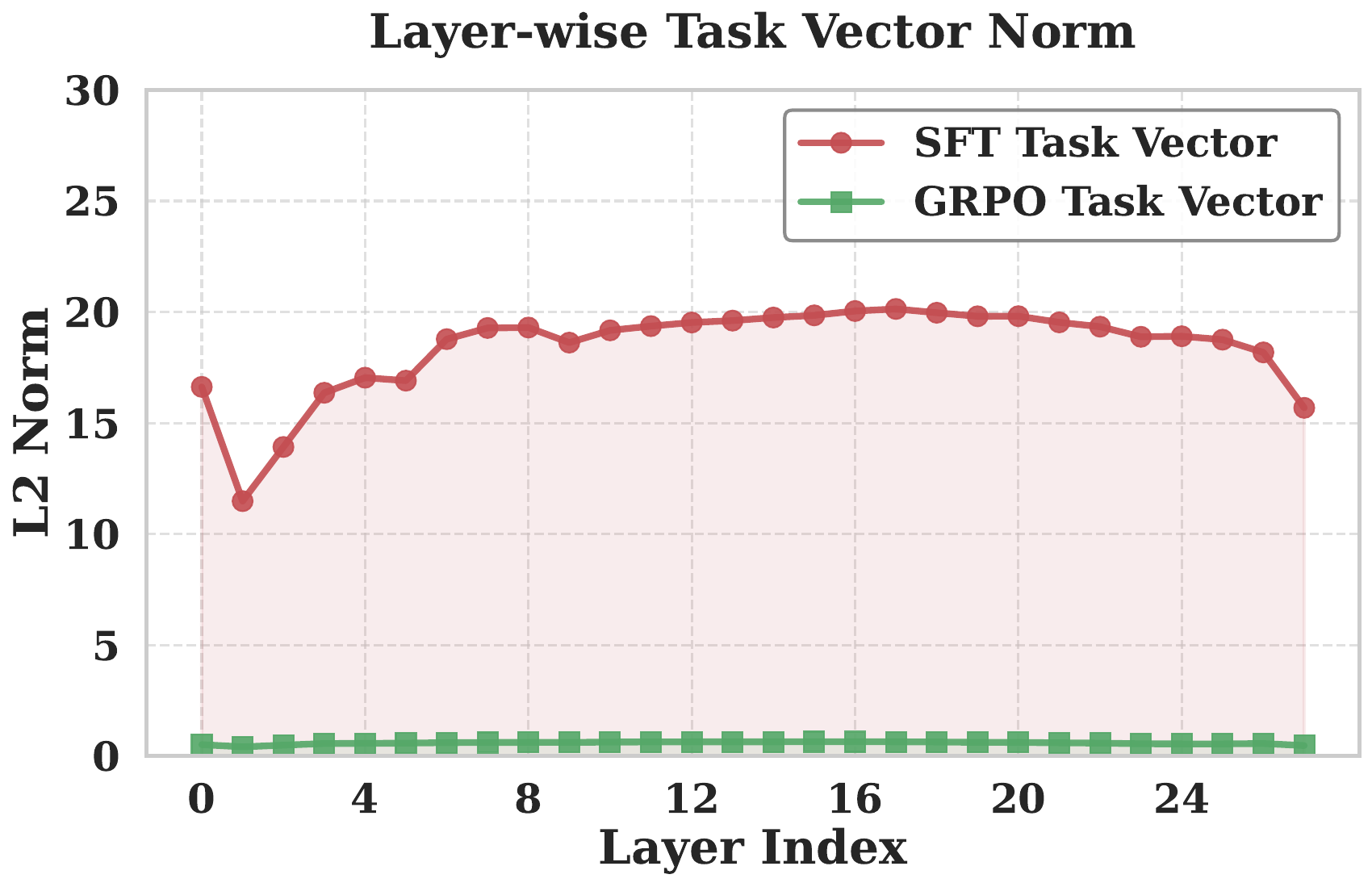}
        \caption{\small Layer-wise L2 norm of each task vector.}
        \label{fig:layerwise_norm}
    \end{subfigure}
    \hfill
    \begin{subfigure}[t]{0.31\textwidth}
        \centering
        \includegraphics[width=\linewidth]{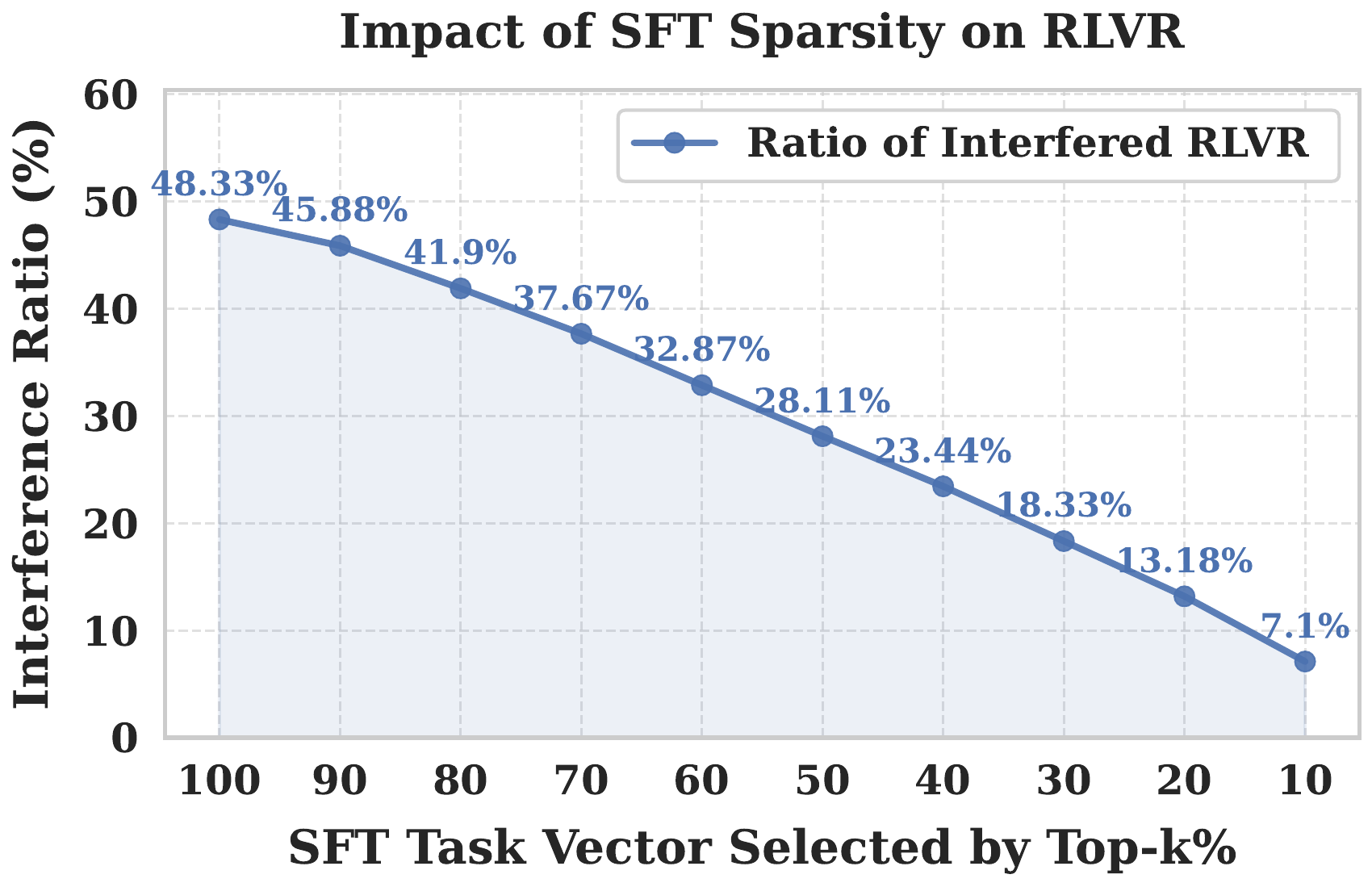}
        \caption{\small Sign interference under different SFT sparsification ratios.}
        \label{fig:sparsity_impact}
    \end{subfigure}
    \hfill
    \begin{subfigure}[t]{0.31\textwidth}
        \centering
        \raisebox{0.3em}
        {\includegraphics[width=\linewidth]{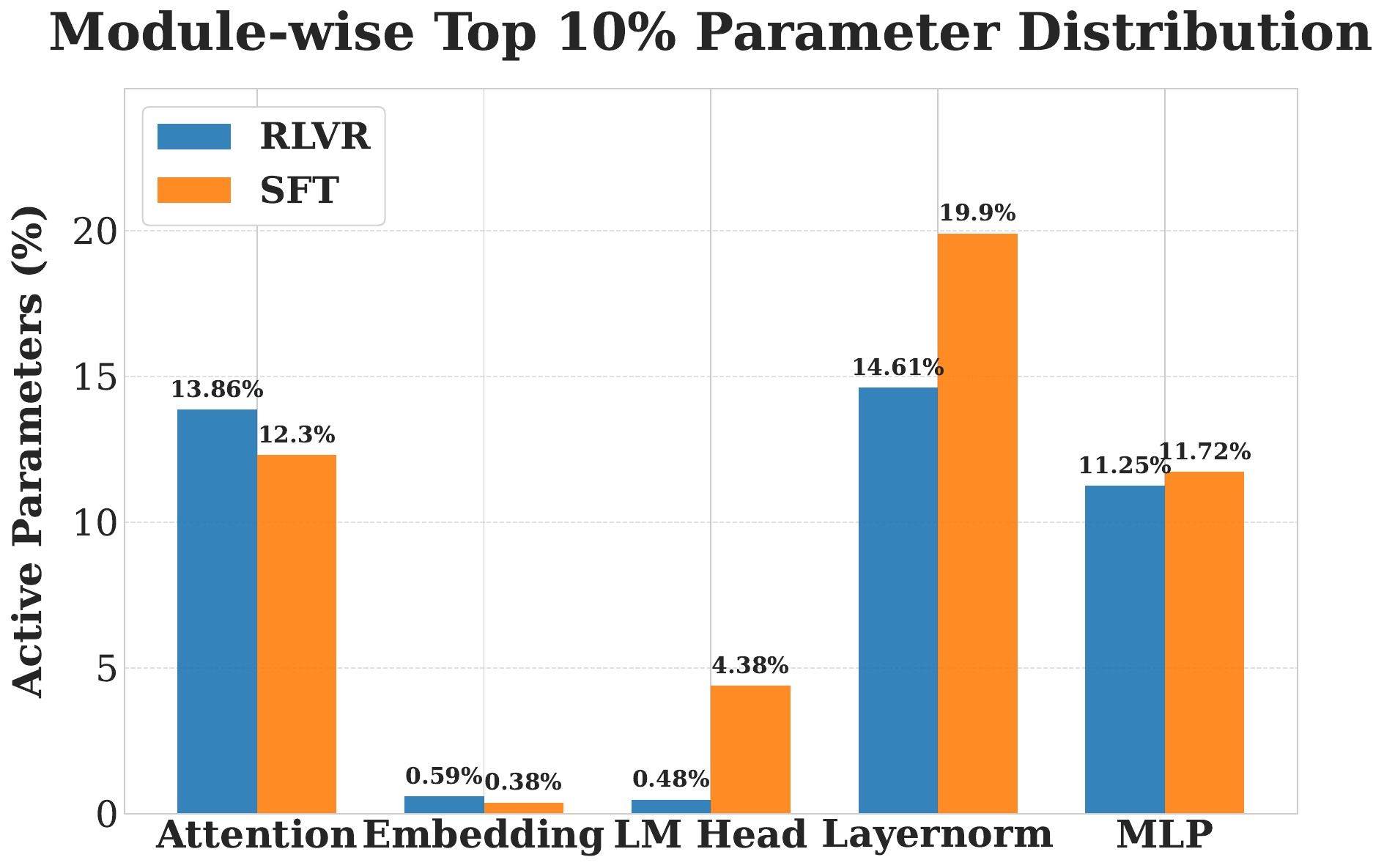}}
        \caption{\small Module-wise distribution of activated parameters.}
        \label{fig:modulewise_activated}
    \end{subfigure}
    \vspace{-1ex}
    \caption{\small Empirical analysis of SFT and RLVR task vectors. 
    \textbf{(a)} Layer-wise L2 norms show an approximately $30\times$ magnitude gap between SFT and RLVR. 
    \textbf{(b)} SFT sparsification substantially reduces sign interference with the sparsified RLVR task vector. The y-axis measures the fraction of parameters with opposite signs among the non-zero entries of the sparsified RLVR task vector. Without sparsification, 44.91\% of parameters exhibit sign conflicts. 
    \textbf{(c)} The top 10\% activated parameters follow different module-wise distributions in SFT and RLVR, suggesting that the two paradigms affect partly complementary components of the model.
    }
    \vspace{-3ex}
    \label{fig:combined_analysis}
\end{figure*}

\section{Understanding SFT and RLVR Task Vectors}\label{sec:analysis}

\paragraph{Preliminary.}
Following the standard formulation~\cite{ilharco2023editing}, given the parameters $\theta$ of a pre-trained base model, a \textit{task vector} is defined as the element-wise difference between post-trained parameters $\theta_{\text{ft}}$ and the base model: $\tau = \theta_{\text{ft}} - \theta$.
In our setting, SFT and RLVR are applied independently to the same base model $\theta$ and training dataset, yielding two task vectors:
\begin{equation}
    \tau_{\text{SFT}} = \theta_{\text{SFT}} - \theta, \qquad
    \tau_{\text{RLVR}} = \theta_{\text{RLVR}} - \theta.
\end{equation}
A merged model can then be constructed as $\theta_{\text{merged}} = \theta + \lambda_{\text{SFT}} \cdot \tau_{\text{SFT}} + \lambda_{\text{RLVR}} \cdot \tau_{\text{RLVR}}$ where $\lambda_{\text{SFT}}$ and $\lambda_{\text{RLVR}}$ are scalar coefficients.
We analyze these task vectors on Qwen2.5-Math-7B~\cite{yang2024qwen2} and identify three structural properties. Together, they help explain why direct integration is difficult and motivate the design of \ours.

\subsection{Finding 1: Extreme Magnitude Disparity}\label{sec:finding1}

We first examine the layer-wise magnitudes of the SFT and RLVR task vectors. 
As shown in Fig.~\ref{fig:layerwise_norm}, the L2 norm of $\tau_{\text{SFT}}$ is consistently much larger than that of $\tau_{\text{RLVR}}$ across layers. 
The SFT task vector has norms around 15 to 20 in most layers, whereas the RLVR task vector remains below 1. 
Overall, we observe $\|\tau_{\text{SFT}}\|_2 \approx 30 \times \|\tau_{\text{RLVR}}\|_2$. This scale gap is much larger than what is usually assumed in model merging, where task vectors are expected to have more comparable magnitudes.

This imbalance creates a direct problem for task-vector merging. In a naive linear combination, the merged update is dominated by the SFT task vector, and the weaker but important RLVR reasoning signal can be overwhelmed. This helps explain why standard merging methods such as TIES-Merging~\cite{yadav2023tiesmerging} and DARE~\cite{yu2024language} perform poorly in our setting, as shown in Tab.~\ref{tab:merged_results}. Importantly, the smaller norm of the RLVR task vector should not be interpreted as lower importance. Prior work suggests that RLVR updates can be nearly orthogonal to pretrained weights~\cite{zhu2025path}, indicating a different update geometry that direct merging must account for.

\textbf{Implication for \ours.}
The magnitude gap motivates a conflict-aware sparsification strategy before merging. Instead of allowing the large SFT task vector to dominate the entire merged update, we retain only the most important entries of each task vector and rescale them before coefficient search. This reduces unnecessary interference and creates a more balanced basis for combining SFT and RLVR capabilities.

\subsection{Finding 2: Severe Sign Interference}\label{sec:finding2}

Beyond magnitude differences, we next examine directional conflicts between $\tau_{\text{SFT}}$ and $\tau_{\text{RLVR}}$. We measure \textit{sign interference} as the fraction of parameters where the two task vectors have opposite signs, which indicates conflicting update directions. Without any processing, 44.91\% of parameters show sign disagreement between the SFT and RLVR task vectors. Thus, nearly half of the parameters receive opposing update signals from the two post-training paradigms. Consistently, Tab.~\ref{tab:grad_conflict} shows that the gradient conflict ratio between SFT and RLVR objectives remains close to $0.5$ during joint training. This suggests that the conflict is not merely a transient optimization issue, but a persistent source of difficulty in joint SFT--RLVR training.

This interference is not evenly distributed across parameters. Prior work suggests that fine-tuning knowledge is often concentrated in high-magnitude parameters~\cite{li2025when}, implying that task vectors contain many low-magnitude entries that may contribute more noise than useful signal. We therefore test whether keeping only the most important entries can reduce sign interference. Specifically, we sparsify the SFT task vector at different retention rates and measure its sign conflict with the top-10\% sparsified RLVR task vector.

As shown in Fig.~\ref{fig:sparsity_impact}, sign interference decreases steadily as the SFT task vector becomes sparser. The conflict ratio drops from 48.33\% under full SFT retention to 7.1\% when only the top-10\% SFT parameters are kept. This result indicates that many sign conflicts come from low-magnitude entries, and that magnitude-based pruning can effectively reduce directional interference while retaining the dominant update directions.

\textbf{Implication for \ours.}
Sign interference helps explain why joint training can suffer from persistent gradient conflicts. It also motivates the magnitude-based pruning used in \ours. By removing low-magnitude entries before merging, \ours reduces unnecessary conflicts and preserves the task-specific update directions that are most likely to matter.

\subsection{Finding 3: Heterogeneous Important Parameter Distributions}\label{sec:finding3}

After observing strong conflicts in magnitude and sign, we next ask \textit{whether SFT and RLVR affect the same parts of the model}. We retain the top 10\% parameters by magnitude in each task vector and examine how these activated parameters are distributed across different Transformer modules.

As shown in Fig.~\ref{fig:modulewise_activated}, SFT and RLVR show clearly different module-wise patterns. SFT concentrates its important updates in LayerNorm modules (19.9\%) and has very limited activation in the LM head (0.48\%). In contrast, RLVR distributes its updates more broadly, with substantial activation in attention (13.86\%), LayerNorm (14.61\%), and the LM head (4.38\%). We observe a similar pattern on LLaMA3.1-8B in Appendix~\ref{appendix:cross_architecture}, especially for LayerNorm and the LM head, suggesting that this difference is not specific to Qwen2.5-Math-7B.

This finding provides a positive signal that complements Findings 1 and 2. Although SFT and RLVR conflict in magnitude and sign, their largest updates are not concentrated in exactly the same modules. SFT and RLVR therefore appear to modify partly different components of the model. This suggests that their updates are not purely redundant or purely antagonistic. Instead, they may encode complementary capabilities that can be combined if interference is properly controlled.

\textbf{Implication for \ours.}
Finding 3 supports the feasibility of test-time synthesis. If SFT and RLVR contribute through partly different parameter regions, then a post-hoc composition strategy can preserve useful signals from both source models without forcing the two objectives to interact during training. This motivates the decoupled design of \ours, where SFT and RLVR are trained independently and combined only after conflict-aware processing.

\paragraph{Summary.}
Together, these three findings show that SFT and RLVR are difficult to merge naively because of scale imbalance and sign conflict, but they also contain complementary signals at the module level. The key challenge is therefore not simply whether SFT and RLVR should be combined, but how to combine them while controlling interference. This analysis motivates the design of the decoupled and conflict-aware test-time synthesis framework developed in Sec.~\ref{sec:methodology}.

\section{Related Works}

\subsection{SFT--RLVR: Divergence and Integration} 

Prior work has studied the differences between SFT and RLVR from several perspectives. 
\citet{chu2025sft} show that SFT tends to memorize training distributions, while RL generalizes better. \citet{rajani2025scalpel} and \citet{matsutani2025rl} find that SFT can overwrite pretrained knowledge, whereas RLVR tends to amplify existing capabilities. \citet{zhu2025path} further show that RLVR updates are often close to orthogonal to pretrained weights, while SFT updates are more aligned with them. Our analysis in Sec.~\ref{sec:analysis} complements these studies by quantifying the SFT--RLVR difference at the task-vector level, including magnitude disparity, sign interference, and module-wise update distributions.

Recent methods attempt to combine SFT and RLVR by interleaving the two procedures or jointly optimizing their objectives~\cite{ma2025learning, yan2025learning, zhang2025bread, wu2025thought, fu2025srft, qin2025supervised, wu2025generalization, chen2025bridging}. These methods can be effective, but they require additional training and careful tuning to balance supervised and reinforcement learning signals. In contrast, \ours decouples the two training processes and performs capability synthesis only at test time. This avoids directly optimizing conflicting objectives during post-training.

\subsection{Model Merging}
Task arithmetic~\cite{ilharco2023editing} shows that task vectors from homogeneous fine-tuned models can be linearly composed. Subsequent methods such as TIES-Merging~\cite{yadav2023tiesmerging} and DARE~\cite{yu2024language} improve merging robustness through sparsification and rescaling. Test-time merging methods, including AdaMerging~\cite{yang2024adamerging} and WUDI-Merging~\cite{cheng2025whoever}, further adapt merging coefficients at inference time. However, these methods are mainly designed for task vectors obtained from similar training procedures, where the task vectors often have more comparable scales. As shown in Sec.~\ref{sec:finding1}, SFT and RLVR task vectors violate this assumption because their magnitudes differ by about $30\times$. This scale mismatch, together with sign interference, makes standard merging methods unreliable in our setting, as confirmed by the TIES-Merging and DARE results in Sec.~\ref{sec:experiment}.

\paragraph{Model Merging in LLM Post-training.} 
Model merging has also been explored for LLM alignment and post-training. HMA~\cite{lin2024mitigating} interpolates models to reduce the alignment tax in RLHF. WARP~\cite{rame2024warp} and WARM~\cite{rame2024warm} average policies or reward models to improve robustness along the reward-KL trade-off. These methods usually merge models within the same training paradigm, where the models differ by checkpoint, reward, or training stage. By contrast, \ours aims to combine two structurally different post-training paradigms, SFT and RLVR. This setting requires explicit conflict control before task-vector synthesis, rather than relying on direct interpolation alone.

\section{Methodology}\label{sec:methodology}
\begin{figure*}
    \centering
    \includegraphics[width=0.9\linewidth]{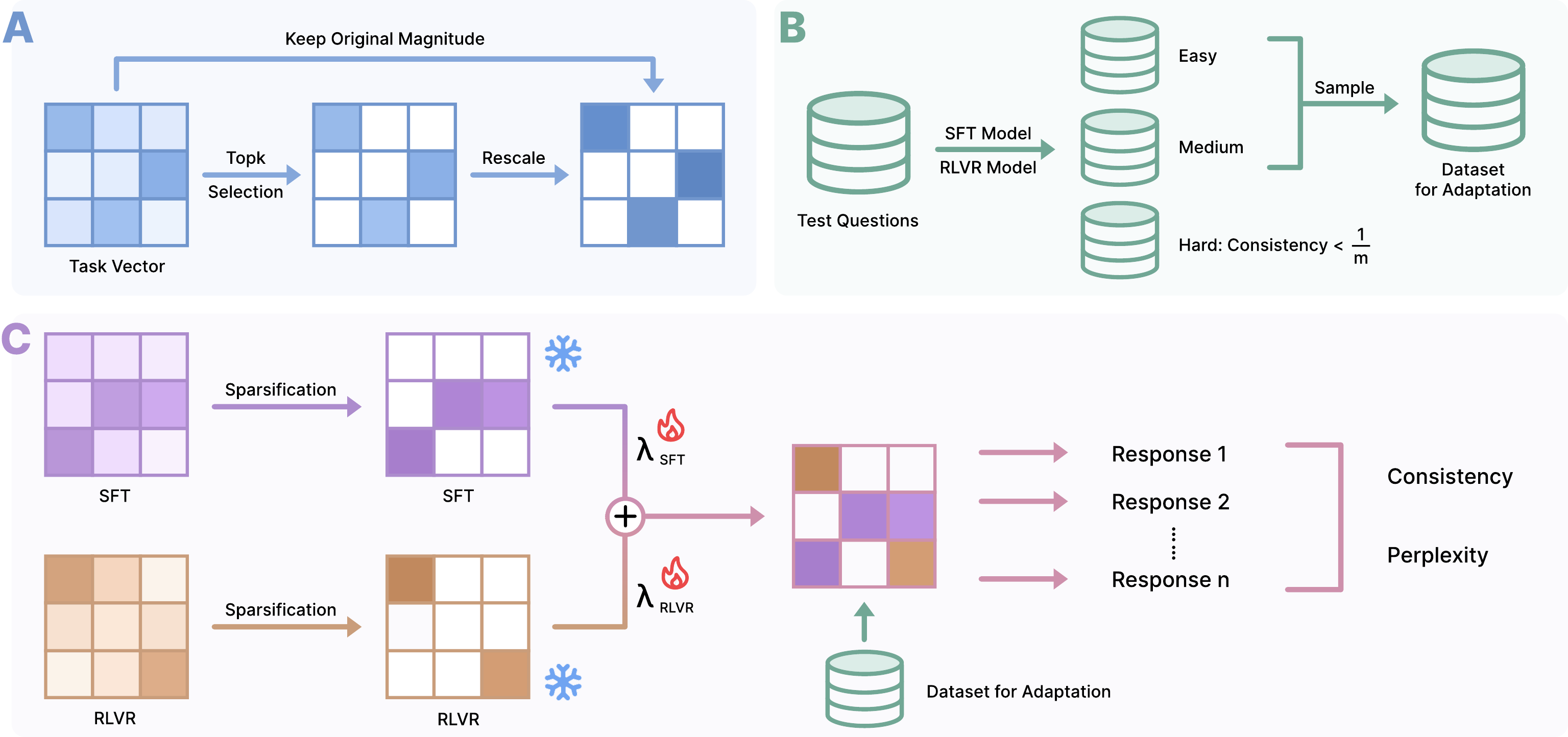}
    \vspace{-1ex}
    \caption{\small Overview of the \ours framework. 
    \textbf{(A) Selective Sparsification.} Each task vector is pruned to retain only its top-$k\%$ entries by magnitude and then rescaled to preserve its original L2 norm.
    \textbf{(B) Difficulty-Aware Data Selection.} A small set of unlabeled queries is stratified by difficulty, estimated from the answer consistency of the SFT and RLVR source models, to provide a reliable signal for coefficient search.
    \textbf{(C) Bayesian Coefficient Optimization.} The sparsified task vectors are combined with scalar weights $\lambda_\text{SFT}$ and $\lambda_\text{RLVR}$. Bayesian optimization searches for coefficients that balance output consistency and perplexity.
     }
     
    \label{fig:ours_framework}
\end{figure*}

The findings in Sec.~\ref{sec:analysis} motivate the design of \textbf{D}ec\textbf{o}upled \textbf{T}est-time \textbf{S}ynthesis (\textbf{\ours}), a three-stage framework for combining SFT and RLVR capabilities without additional post-training. As shown in Fig.~\ref{fig:ours_framework} and Algo.~\ref{alg:dyst}, \ours first sparsifies the SFT and RLVR task vectors, then selects a small set of unlabeled adaptation queries, and finally searches for two scalar combination coefficients.Unlike training-based integration methods, \ours does not update model parameters. It operates directly on existing SFT and RLVR checkpoints, and the only optimization is a lightweight search over $\lambda_\text{SFT}$ and $\lambda_\text{RLVR}$.

\subsection{Selective Sparsification of Task Vectors}\label{subsec:sparsification}

Findings~1 and~2 identify two obstacles to direct task-vector merging. First, the SFT task vector is much larger than the RLVR task vector, so a naive linear combination can be dominated by SFT.
Second, the two task vectors have substantial sign interference, which means that many parameters receive opposing update directions. Finding~2 further shows that this interference can be greatly reduced by keeping only high-magnitude entries. We therefore apply \textit{selective sparsification} to both task vectors before merging.

\paragraph{Magnitude-based pruning.}
Following prior work on task-vector sparsity~\cite{li2025when, yadav2023tiesmerging}, we retain the top-$p\%$ entries by absolute magnitude in each task vector and set all other entries to zero.
Formally, for a task vector $\tau$, we define a binary mask $M$ and obtain
\begin{equation}
    \tilde{\tau}_{\text{sparse}} = \tau \odot M, \quad
    M_i = \mathbbm{1}\bigl[|\tau_i| \geq \text{quantile}_{1-p}(|\tau|)\bigr].
\end{equation}
This removes low-magnitude entries that are more likely to introduce noise or interference, while keeping the dominant task-specific update directions. We use $p\%=30\%$ by default.

\paragraph{Norm-preserving rescaling.}
Pruning reduces the L2 norm of a task vector, which may weaken the effect of the retained update and make coefficient search unstable. We therefore rescale the pruned vector to match the norm of the original task vector:
\begin{equation}
    \tilde{\tau} = \gamma \cdot \tilde{\tau}_{\text{sparse}}, \quad
    \gamma = \frac{\|\tau\|_2}{\|\tilde{\tau}_{\text{sparse}}\|_2 + \epsilon}.
\end{equation}
This step preserves the overall strength of each task vector after pruning. The remaining scale difference between SFT and RLVR is then handled by the learned coefficients in Sec.~\ref{subsec:bayesian}.

\subsection{Difficulty-Aware Data Selection}
\label{subsec:dataselection}

Coefficient optimization requires a small set of unlabeled queries that can distinguish good coefficient choices from poor ones. 
Randomly sampled queries may be uninformative. Very easy queries are often answered consistently by both source models, while very hard queries may be unstable for both models.
We therefore construct a \textit{difficulty-stratified adaptation set} using only unlabeled queries and no ground-truth answers.

\paragraph{Difficulty scoring.}
For each candidate query $x$, we generate $m=5$ inference paths from both the SFT model $\theta_\text{SFT}$ and the RLVR model $\theta_\text{RLVR}$.
We measure answer consistency by majority vote:
\begin{equation}\label{eq:consistency}
    C_\theta(x) = \max_{y \in \mathcal{Y}} \, \frac{1}{m} \sum_{i=1}^m \mathbb{I}(f_\theta(x)_i = y).
\end{equation}
The difficulty score of $x$ is defined as
\begin{equation}
    D(x) = 1 - \frac{1}{2} \left( C_{\theta_\text{SFT}}(x) + C_{\theta_\text{RLVR}}(x) \right).
\end{equation}
A larger $D(x)$ indicates that both source models produce less consistent answers, while a smaller $D(x)$ indicates that both models are more stable.

\paragraph{Stratified sampling.}
We discard queries with $D(x)>1-\frac{1}{m}$, since these queries are too unstable to provide reliable optimization signals. The remaining queries are split at the median into low-difficulty and medium-difficulty pools. We then sample equally from the two pools to form the adaptation set.
By default, we use 64 queries.
Low-difficulty queries provide stable signals from at least one source model, while medium-difficulty queries help distinguish coefficient settings that better combine SFT and RLVR capabilities. The ablation study in Sec.~\ref{subsec:ablation} confirms that this strategy outperforms random sampling and difficulty-uniform sampling.

\subsection{Bayesian Coefficient Optimization}
\label{subsec:bayesian}

Given the sparsified task vectors $\tilde{\tau}_\text{SFT}$ and $\tilde{\tau}_\text{RLVR}$, we define the merged model as
\begin{equation}
    \theta = \theta_\text{base}
    + \lambda_\text{SFT} \tilde{\tau}_{\text{SFT}}
    + \lambda_{\text{RLVR}} \tilde{\tau}_{\text{RLVR}},
    \label{eq:merge}
\end{equation}
where $\theta_\text{base}$ is the base model, and $\lambda_\text{SFT}, \lambda_\text{RLVR} \in [0,2]$ are the coefficients to be optimized. Fixed symmetric choices such as $(1,1)$ or $(0.5,0.5)$ cannot reliably account for the different strengths and effects of the two task vectors. We therefore search for the coefficients adaptively.

\paragraph{Primary objective.}
For each candidate pair $(\lambda_\text{SFT}, \lambda_\text{RLVR})$, we construct the merged model using Eq.~\eqref{eq:merge}. We then generate $k$ outputs for each adaptation query and compute the average consistency score using Eq.~\eqref{eq:consistency}. Higher consistency indicates that the merged model produces more stable answers. Although consistency is not a direct measure of correctness, it provides a useful label-free signal for coefficient selection.

\paragraph{Secondary constraint.}
Maximizing consistency alone can lead to degenerate solutions, since a model that repeatedly produces the same invalid output may also appear highly consistent. We therefore use perplexity on the adaptation queries as a secondary criterion. Coefficient selection is performed on the Pareto frontier that maximizes consistency while keeping perplexity low.

\paragraph{Search algorithm and model selection.}
Since consistency is non-differentiable with respect to the coefficients, we use the Tree-structured Parzen Estimator (TPE)~\cite{bergstra2011algorithms} for black-box Bayesian optimization. We run 100 trials in the two-dimensional coefficient space. For stronger backbones such as the Qwen2.5-Math series, perplexity remains stable during coefficient search, so we select the Pareto point with the highest consistency. For weaker backbones such as LLaMA3.1-8B, high consistency can coincide with degraded perplexity, so we select the knee point on the Pareto frontier. Further details and visualizations are provided in Appendix~\ref{appendix:pareto_front}.

\begin{table*}[t]
\centering
\caption{\small Overall performance on \textbf{Qwen2.5-Math-7B}. We compare foundation models, prior RLVR methods, training-based SFT--RLVR integration methods, model merging baselines, and \ours.
}
\vspace{-1.5ex}
\label{tab:merged_results}
\setlength{\tabcolsep}{10pt}  
\renewcommand{\arraystretch}{0.95} 
\resizebox{\textwidth}{!}{%
\begin{tabular}{lccccccc}
\toprule
  \textbf{Model} & \textbf{AIME 24} & \textbf{AIME 25} &\textbf{AMC} & \textbf{MATH500} & \textbf{Minerva} & \textbf{Olympiad} & \textbf{Average} \\
\midrule
Qwen2.5-Math-7B-base~\cite{yang2024qwen2}       
  & 11.5&4.9    & 31.3 & 43.6 & 7.4  & 15.6 & 19.0      \\
Qwen2.5-Instruct-7B~\cite{yang2024qwen2} 
  & 12.5&10.2   & 48.5 & 80.4 & 32.7 & 41.0 & 37.6      \\
\midrule
\multicolumn{8}{c}{Previous RLVR methods} \\
\midrule
SimpleRL-Zero~\cite{zeng2025simplerlzoo}                 
  & 27.0&6.8    & 54.9 & 76.0 & 25.0 & 34.7 & 37.4      \\
OpenReasoner-Zero~\cite{hu2025openreasonerzero}                  
  & 16.5&15.0   & 52.1 & 82.4 & 33.1 & 47.1 & 41.0      \\
PRIME-Zero~\cite{cui2025process}                       
  & 17.0&12.8   & 54.0 & 81.4 & 39.0 & 40.3 & 40.7      \\
Oat-Zero~\cite{liu2025understanding}                        
  & 33.4 &11.9 & 61.2 & 78.0 & 34.6 & 43.4 & 43.7      \\
\midrule
\multicolumn{8}{c}{SFT, RLVR and Their Combined Models} \\
\midrule
SFT                                           
  & 22.2&22.3 & 52.8 & 82.6 & 40.8  & 43.7 & 44.1      \\
On-Policy RL                                  
  & 25.1&15.3   & 62.0 & 84.4 & 39.3 & 46.8 & 45.5 \\

RL w/ SFT Loss &19.5&16.4&  49.7&   80.4&   34.9&   39.4& 40.1 \\
SFT+RL & 25.8 &23.1  & 62.7 & 87.2 & 39.7 & 50.4 &48.2 \\
LUFFY~\cite{yan2025learning} & 27.1 & 22.3 & 64.6 & 86.8 & 35.7 & 58.5 & 49.2 \\
ExGRPO~\cite{zhan2026exgrpo} & 28.5 & 17.8 & 65.7 & 86.0 & 38.2 & 51.0 & 47.9 \\
ReLIFT~\cite{ma2025learning} & 26.4 & 19.7 & 64.4 & 86.2 & 32.7 & 53.6 & 47.9 \\
\midrule
\multicolumn{8}{c}{Model Merging Baselines} \\
\midrule
TIES-Merging (SFT + On-Policy RL)   & 15.9 & 9.4 & 51.4 &  75.8 & 33.8 & 40.7 & 37.8  \\
TIES-Merging$^*$ (SFT + On-Policy RL)   & 27.6 & 21.4 & 62.3 &  87.6 & 42.3 & 49.0 & 48.4  \\
DARE (SFT + On-Policy RL) & 6.8 & 6.1 & 16.1 & 34.4 & 8.5 & 14.8 & 14.5   \\
\midrule
\multicolumn{8}{c}{\textbf{Our Methods}} \\
\midrule
\textbf{\ours (SFT + On-Policy RL)}                                      
  & 32.9& 23.8 & 63.2 & 86.8 & 42.6 & 48.3 & \textbf{49.3} \\
\textbf{\ours (ExGRPO + ReLIFT)} & 31.8 & 22.9 & 66.1 & 87.4 & 37.9 & 57.8 & \textbf{50.6} \\

\bottomrule
\end{tabular}%
}
\end{table*}

\section{Experiment}
\label{sec:experiment}

We evaluate \ours on mathematical reasoning benchmarks across different model backbones. Our main goal is to test whether test-time task-vector synthesis can match or exceed training-based SFT--RLVR integration while using substantially less computation.

\subsection{Experimental Setting}
\label{sec:main_exp_setting}

\paragraph{Source checkpoints and task vectors.} We use Qwen2.5-Math-7B~\cite{yang2024qwen2} as the primary backbone. The base model, SFT checkpoint, and On-Policy RL checkpoint are taken from prior work~\cite{yan2025learning}\footnote{\url{https://huggingface.co/collections/Elliott/luffy-rl}}. All these are post-trained on a subset of OpenR1-Math-220k~\cite{face2025open}. \ours directly extracts task vectors from these public checkpoints and does not require additional training.

\paragraph{Evaluation benchmarks.}
We evaluate on six mathematical reasoning benchmarks: AIME 2024, AIME 2025, AMC~\cite{li2024numinamath}, MATH500~\cite{hendrycks2021measuring}, Minerva, and OlympiadBench~\cite{he2024olympiadbench}. For benchmarks with limited test samples, including AIME 2024, AIME 2025, and AMC, we report average@32. For MATH500, Minerva, and OlympiadBench, we report pass@1. Following prior work~\cite{yan2025learning}, we use temperature 0.6 and set the maximum generation length to 8192 tokens for all evaluations.

\paragraph{Baselines.} 
We compare with four groups of baselines. 
\textit{Foundation models} include Qwen2.5-Math-7B-base and Qwen2.5-Math-7B-Instruct~\cite{yang2024qwen2}. 
\textit{Training-based SFT--RLVR integration methods} include SFT, On-Policy RL, RL w/ SFT Loss, SFT+RL, LUFFY~\cite{yan2025learning} and ReLIFT~\cite{ma2025learning}. SFT and On-Policy RL are also the source models used by \ours for task-vector extraction. 
\textit{RLVR-only methods trained from the base model} include SimpleRL-Zero~\cite{zeng2025simplerlzoo}, OpenReasoner-Zero~\cite{hu2025openreasonerzero}, PRIME-Zero~\cite{cui2025process}, Oat-Zero~\cite{liu2025understanding} and ExGRPO~\cite{zhan2026exgrpo}. 
\textit{Model merging baselines} include TIES-Merging~\cite{yadav2023tiesmerging} and DARE~\cite{yu2024language}, using the same source models as \ours. 
We also include TIES-Merging$^*$, which replaces the original TIES sparsification strategy with our sparsification module.

\subsection{Main Results on Qwen2.5-Math-7B}
\label{sec:main_results}

Tab.~\ref{tab:merged_results} reports the main results on Qwen2.5-Math-7B. We discuss the results from three perspectives.

\paragraph{Why do standard merging methods fail?}
The model merging baselines support the analysis in Sec.~\ref{sec:analysis}. DARE obtains an average score of 14.5, the lowest result in the table. One likely reason is that DARE assumes task-vector updates are sparse and low-magnitude, an assumption that does not hold for the SFT task vector in this setting. TIES-Merging also performs poorly, with an average score of 37.8. 
Its sign-election mechanism is designed for task vectors with more comparable scales, but the large SFT--RLVR magnitude gap can cause the SFT direction to dominate the merged update and weaken the RLVR contribution. TIES-Merging$^*$ improves to 48.4 after replacing its sparsification mechanism with ours, but it still falls below \ours. This suggests that sparsification is useful but not sufficient. Adaptive coefficient search is also needed.

\paragraph{How does \ours compare to training-based integration?}
\ours (SFT + On-Policy RL) achieves an average score of 49.3. It outperforms both source models, SFT at 44.1 and On-Policy RL at 45.5. It also matches or exceeds training-based integration methods, including RL w/ SFT Loss at 40.1, SFT+RL at 48.2, and LUFFY at 49.2. 
The comparison with SFT+RL is especially relevant because sequential training is a common way to combine SFT and RLVR. \ours improves over this baseline without any additional post-training, supporting the benefit of decoupling training from test-time synthesis.

\paragraph{Does \ours work with stronger checkpoints?} 
To test whether \ours is limited to vanilla SFT and RLVR, we apply it to ExGRPO~\cite{zhan2026exgrpo}, a stronger RLVR model, and ReLIFT~\cite{ma2025learning}, a stronger post-trained checkpoint that already encode complementary knowledge and reasoning behaviors on hard questions.
As shown in Tab.~\ref{tab:merged_results}, \ours (ExGRPO + ReLIFT) achieves an average of 50.6.
It improves over both ExGRPO and ReLIFT by 2.7 points and outperforms LUFFY by 1.4 points. This indicates that \ours can serve as a general post-hoc composition method, rather than only a remedy for weak source checkpoints.

\subsection{Training Efficiency}
\label{sec:efficiency}

\begin{table}[tb]
    \centering
    \vspace{-1.5ex}
    \caption{Comparison of computational costs and data requirements.}
    \vspace{-1ex}
    \renewcommand{\arraystretch}{0.95} 
    \resizebox{0.8\linewidth}{!}{
    \begin{tabular}{l l l}
    \toprule
    \textbf{Model} & \textbf{GPU Hours} & \textbf{Data Usage~(On/Off)} \\\midrule
    \textbf{LUFFY} & 77 $\times$ 8 & ~~64K $\times$ 7~/~64K \\
    \textbf{SFT} & 24 $\times$ 8 & \quad~~~~0\quad~~~/~~64K \\
    \textbf{RL w/ SFT Loss}& 133 $\times$ 8 & ~~64K $\times$ 7~/~64K \\
    \textbf{SFT+RL} & 130 $\times$ 8 & ~~64K $\times$ 8~/~135K \\\midrule
    \textbf{\ours} & \textbf{10 $\times$ 2} & \textbf{\quad~~~~~64\quad~/~~~0} \\\bottomrule
    \end{tabular}
    }
    \label{tab:comparison_resources}
\end{table}

Tab.~\ref{tab:comparison_resources} compares the computational cost and data usage of \ours with training-based baselines. Because \ours operates on existing public checkpoints, its cost comes only from Bayesian coefficient search. It uses 20 GPU hours in total, compared with 616 GPU hours for LUFFY and more than 1,000 GPU hours for RL w/ SFT Loss and SFT+RL. This corresponds to about 3\% of the training cost of these integration methods. \ours also requires only 64 unlabeled queries and no off-policy data, while training-based methods use tens of thousands of on-policy and off-policy samples. As shown in Sec.~\ref{sec:main_results}, this reduction in computation and data does not come at the cost of performance.

\begin{table}[t]
\centering
\caption{\small OOD QA performance with strong source checkpoints. Coefficients are learned on mathematical reasoning samples and directly transferred without re-tuning.}
\vspace{-1.5ex}
\label{tab:ood_results}
\setlength{\tabcolsep}{6pt}
\renewcommand{\arraystretch}{0.95}
\resizebox{0.99\linewidth}{!}{%
\begin{tabular}{lcccc}
\toprule
\textbf{Model} & \textbf{ARC-C} & \textbf{GPQA} & \textbf{MMLU-Pro} & \textbf{Average} \\
\midrule
LUFFY & 80.5 & 39.9 & 53.0 & 57.8 \\
ExGRPO~\cite{zhan2026exgrpo} & \textbf{84.7} & 37.4 & 52.9 & 58.3 \\
ReLIFT~\cite{ma2025learning} & 82.8 & 36.9 & 53.9 & 57.9 \\
\textbf{\ours (ExGRPO + ReLIFT)} & 84.6 & \textbf{42.9} & \textbf{53.9} & \textbf{60.5} \\
\bottomrule
\end{tabular}%
}
\vspace{-3ex}
\end{table}

\begin{table*}[t]
\centering
\caption{Overall performance based on alternative backbones, including \textbf{Qwen2.5-Math-1.5B} and \textbf{LLaMA3.1-8B}. All models are evaluated under a unified setting. 
}
\vspace{-1.5ex}
\label{tab:qwen15_results}
\setlength{\tabcolsep}{10pt}  
\renewcommand{\arraystretch}{0.90} 
\resizebox{\textwidth}{!}{%
\begin{tabular}{lccccccc}
\toprule
  \textbf{Model} & \textbf{AIME 24} & \textbf{AIME 25} &\textbf{AMC} & \textbf{MATH500} & \textbf{Minerva} & \textbf{Olympiad} & \textbf{Average} \\
\midrule
Qwen2.5-Math-1.5B-base~\cite{yang2024qwen2}       
  &  6.0 & 3.9   &  25.7  & 28.6  & 9.9  & 17.8 & 15.3      \\
Qwen2.5-Math-1.5B-Instruct~\cite{yang2024qwen2} 
  & 12.1 & 8.9   & 48.1 & 77.4 & 28.7 & 39.1 & 35.7      \\
  \midrule
\multicolumn{7}{c}{Qwen2.5-Math-1.5B: Training Configuration from DFT~\cite{wu2025generalization}} \\
\midrule
SFT      & 
  3.9& 1.1  & 28.4   & 58.4 & 23.5  & 25.8  & 23.5        \\
DPO~\cite{rafailov2023direct}       
& 9.9  & 5.1   & 40.7 & 65.6 & 16.5  & 30.9  & 28.1      \\
DFT~\cite{wu2025generalization}       
  & 5.7  & 3.2   & 30.3 & 63.0 & 22.8  & 27.9  & 25.5      \\
\textbf{\ours (SFT + DPO)} & 9.3  & 5.4 & 38.8  & 66.8 & 30.5 &  34.8 & \textbf{30.3}  \\       
\midrule
\multicolumn{7}{c}{Qwen2.5-Math-1.5B: Training Configuration from LUFFY~\cite{yan2025learning}} \\
\midrule
SFT & 14.3 & 15.7 & 45.2 & 76.2 & 28.7 & 35.1 & 35.9 \\
On-Policy RL & 10.8 & 7.1 & 45.0 & 74.4 & 28.3 & 38.7 & 34.0 \\
LUFFY~\cite{yan2025learning} & 16.0 & 13.1 & 47.1 & 80.2 & 30.5 & 41.0 & 38.0 \\
\textbf{\ours (SFT + On-Policy RL)} & 17.4 & 18.8 & 50.4 & 79.8 & 29.0 & 44.9 & \textbf{40.0} \\
\midrule
\multicolumn{7}{c}{LLaMA3.1-8B: Training Configuration from LUFFY~\cite{yan2025learning}} \\
\midrule
LLaMA3.1-8B-base~\cite{grattafiori2024llama} & 0.3  & 0.0 & 3.9 & 14.4 & 6.3 & 2.5  & 4.6 \\
LLaMA3.1-8B-Instruct~\cite{grattafiori2024llama} & 5.1& 0.4 & 18.6 & 44.6 & 19.5 &  14.1 & 17.1\\
SFT & 0.6 & 0.2 & 9.4 & 25.8 & 4.4 & 6.7 &  7.9 \\
SFT++ & 1.3 & 7.1 & 21.7 & 45.6 & 9.6 & 18.4 & 17.3 \\
On-Policy RL & 0.0 & 0.0 & 0.8 & 22.8 & 14.0 & 6.5 & 8.6 \\
LUFFY & 1.9 & 0.1 & 13.5 & 39.0 & 15.1 & 9.6 & 13.2 \\
\textbf{\ours(SFT + On-Policy RL)} & 1.0 & 0.3 & 10.3 & 30.4 & 10.3 & 9.8  & 9.9 \\
\textbf{\ours(SFT++ + On-Policy RL)} & 3.3 & 8.3 & 31.2 & 55.6 & 12.5 & 24.4 & \textbf{22.6} \\
\bottomrule
\end{tabular}%
}
\end{table*}

\subsection{OOD Generalization}
\label{sec:ood}

A key question is whether the coefficients found on mathematical reasoning queries overfit to that domain. To test this, we evaluate the same merged model from Sec.~\ref{sec:main_results} on three out-of-domain QA benchmarks: ARC-C, GPQA, and MMLU-Pro. The coefficients are learned only from mathematical reasoning queries and are not re-tuned for these QA benchmarks.

As shown in Tab.~\ref{tab:ood_results}, \ours (ExGRPO + ReLIFT) achieves an average score of 60.5. It outperforms LUFFY by 2.7 points, ExGRPO by 2.2 points, and ReLIFT by 2.6 points. The largest improvement appears on GPQA, where \ours improves over ExGRPO by 5.5 points. These results suggest that the learned coefficients are not merely a math-specific heuristic. They can transfer to broader QA tasks without additional tuning.

\begin{table*}[t]
\centering
\caption{Ablation Study on Qwen2.5-Math-7B. We remove or replace individual components of \ours.}
\vspace{-1.5ex}
\label{tab:ablation_study}
\setlength{\tabcolsep}{10pt}  
\renewcommand{\arraystretch}{0.90} 
\resizebox{0.95\textwidth}{!}{%
\begin{tabular}{lccccccc}
\toprule
  \textbf{Model} & \textbf{AIME 24} & \textbf{AIME 25} &\textbf{AMC} & \textbf{MATH500} & \textbf{Minerva} & \textbf{Olympiad} & \textbf{Average} \\
\midrule
\textbf{\ours}                                       
  & 32.9& 23.8 & 63.2 & 86.8 & 42.6 & 48.3 & \textbf{49.3} \\
  \midrule
\textbf{\ours w/o Data Selection } & 27.4 & 19.9 & 59.9 & 86.2 & 41.9 & 52.7 & 48.0  \\
\textbf{\ours w/o Task Vectors Sparsification} & 30.1 & 22.9 &  61.8 & 87.6 & 43.0 & 49.2 & 49.1  \\
\textbf{\ours with Fixed Coefficients ($1.0, 1.0$)} & 29.4 & 23.9 & 60.7 & 85.4 & 40.4 & 47.3 & 47.8 \\
\textbf{\ours with Fixed Coefficients ($0.5, 0.5$)} & 22.5 & 17.0 & 59.1 & 85.2 & 43.8 & 50.5 & 46.3   \\
\textbf{\ours with Fixed Coefficients ($\frac{1}{\sqrt{2}}$, $\frac{1}{\sqrt{2}}$)} & 28.6 & 22.1 & 62.7 &  85.8 & 43.0 & 48.4 & 48.5  \\
\bottomrule
\end{tabular}%
}
\vspace{-2ex}
\end{table*}

\subsection{Extension to Alternative backbones}

\paragraph{Setup.}
To test generalization across model scales and capabilities, we extend \ours to Qwen2.5-Math-1.5B~\cite{yang2024qwen2} and LLaMA3.1-8B~\cite{grattafiori2024llama}. 
For Qwen2.5-Math-1.5B, we evaluate two training configurations, including the DFT setting~\cite{wu2025generalization} using SFT and DPO checkpoints trained on NuminaMath-CoT, and the LUFFY setting~\cite{yan2025learning} using SFT and On-Policy RL checkpoints trained on OpenR1-Math-220k. 
For LLaMA3.1-8B, we follow the LUFFY training configuration. Since LLaMA3.1-8B is a weaker backbone, prior work filters out hard samples with generation length greater than 2048 tokens to stabilize training. We additionally train an SFT++ checkpoint that retains these harder samples, providing a stronger and more complementary SFT source model for \ours.

\paragraph{Results on Qwen2.5-Math-1.5B.}
Tab.~\ref{tab:qwen15_results} shows that \ours consistently improves over its source models across both Qwen2.5-Math-1.5B settings. In the DFT setting, \ours (SFT + DPO) achieves an average score of 30.3, outperforming SFT, DPO, and DFT. In the LUFFY setting, \ours (SFT + On-Policy RL) reaches 40.0, outperforming both source models and LUFFY. These results show that \ours remains effective at a smaller model scale and across different post-training configurations.

\paragraph{Results on LLaMA3.1-8B.}
The LLaMA3.1-8B results show that the quality and complementarity of source models are important for \ours. When using the standard SFT checkpoint, \ours (SFT + On-Policy RL) reaches 9.9, giving only a modest improvement over On-Policy RL at 8.6. This is likely because the standard SFT checkpoint is trained on filtered, easier data and provides limited complementary knowledge to RLVR. In contrast, \ours (SFT++ + On-Policy RL) reaches 22.6, outperforming LUFFY by 9.4 points. This suggests that \ours benefits more when the source checkpoints encode genuinely complementary capabilities. When SFT contains knowledge from harder samples, \ours can synthesize this knowledge with RLVR reasoning even on a weaker backbone. This also supports the motivation for decoupled synthesis, since directly injecting harder samples into RLVR training can make training less stable.

\begin{figure}
    \centering
    \includegraphics[width=0.65\linewidth]{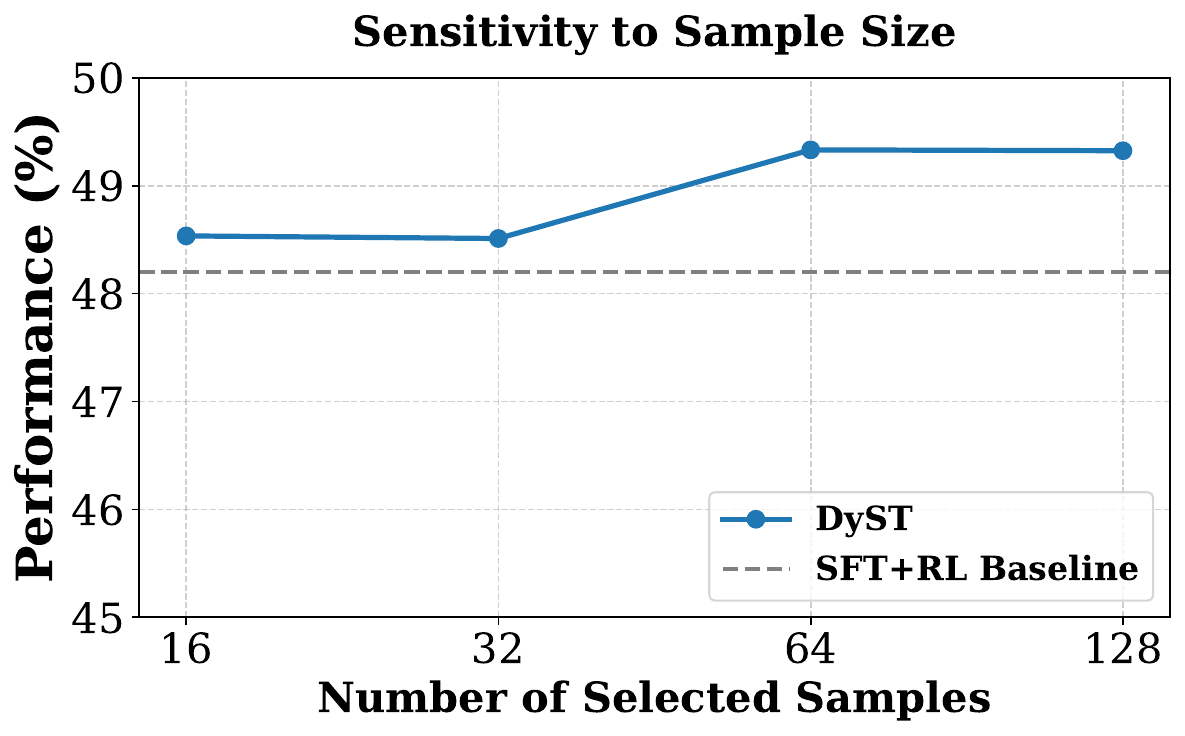}
    \vspace{-1ex}
    \caption{\small Performance vs.\ number of adaptation samples. \ours surpasses the SFT+RL sequential baseline with 16 unlabeled queries, and performance stabilizes around 64 queries.}
    \vspace{-4ex}
    \label{fig:number_of_sample}
\end{figure}

\subsection{Ablation Study}\label{subsec:ablation}

We conduct ablation studies on Qwen2.5-Math-7B to evaluate the contribution of each component in \ours. The results are shown in Tab.~\ref{tab:ablation_study}.

\paragraph{Data selection.}
Replacing difficulty-stratified selection with random sampling reduces the average score from 49.3 to 48.0. This shows that the choice of adaptation queries affects the quality of Bayesian coefficient search. Random sampling can include many queries that are either too easy or too hard, making them less useful for distinguishing coefficient configurations.

\paragraph{Sensitivity to sample size.}
Fig.~\ref{fig:number_of_sample} shows that \ours already surpasses the SFT+RL baseline with 16 unlabeled queries. Performance becomes stable around 64 queries, which we use as the default in all experiments.

\paragraph{Task-vector sparsification.}
Removing sparsification reduces the average score from 49.3 to 49.1. Although the average drop is modest, the effect is clearer on AIME 2024, where performance decreases from 32.9 to 30.1. We further show in Tab.~\ref{tab:sparsification_fixed} that under the same fixed coefficient setting $(1.0,1.0)$, sparsification improves the average score by 0.9 points. This indicates that the benefit of sparsification is not fully absorbed by coefficient tuning.

\paragraph{Bayesian coefficient search.}
Replacing Bayesian search with fixed symmetric coefficients consistently hurts performance. The fixed choices $(1.0,1.0)$, $(0.5,0.5)$, and {\small $\left(\frac{1}{\sqrt{2}}, \frac{1}{\sqrt{2}}\right)$} obtain average scores of 47.8, 46.3, and 48.5, respectively, all below the full \ours result of 49.3. This supports the need for adaptive coefficient selection. The Bayesian search resolves this with small additional cost, using only 100 TPE trials over 64 unlabeled queries.
\section{Conclusion}

We study why integrating SFT and RLVR is difficult in LLM post-training. Through a task-vector analysis, we identify three structural sources of incompatibility: a ${\sim}30{\times}$ magnitude disparity, ${\sim}45\%$ sign interference, and heterogeneous module-wise update distributions. These findings help explain why direct merging and training-based integration can struggle, while also showing that SFT and RLVR encode partly complementary capabilities.

Motivated by this analysis, we propose \textbf{\ours}, a decoupled test-time synthesis framework that combines independently trained SFT and RLVR checkpoints through conflict-aware task-vector composition. \ours uses selective sparsification with norm rescaling to reduce interference, and Bayesian optimization over a small set of unlabeled queries to select combination coefficients. 
Across mathematical reasoning and out-of-domain QA benchmarks, \ours matches or exceeds training-based integration methods while requiring only about 3\% of computational cost. These results suggest that test-time synthesis is an efficient alternative to training-time integration for heterogeneous post-training paradigms.

\paragraph{Limitations and future work.}
\ours depends on the quality and complementarity of its source checkpoints.  When one source model provides limited useful knowledge, as observed on weaker backbones, the benefit of synthesis can be constrained. Future work may explore richer composition strategies, such as layer-wise or module-wise coefficients, and extend the task-vector analysis to broader settings such as open-ended generation and tool use.

\newpage

\bibliography{ref}
\bibliographystyle{icml2026}

\newpage
\appendix
\onecolumn

\begin{figure*}[t]
    \centering
    \begin{subfigure}[t]{0.49\textwidth}
        \centering
        \includegraphics[width=\linewidth]{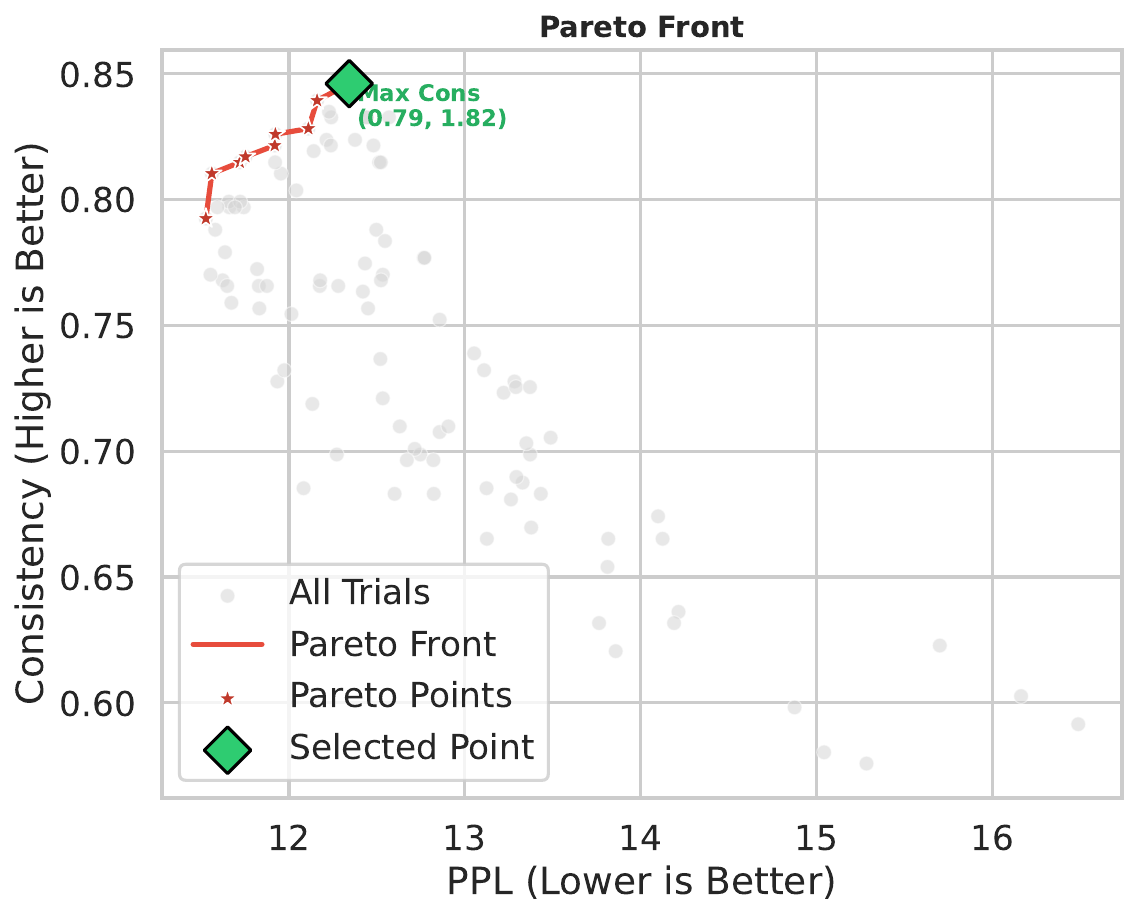}
        \caption{\small Qwen2.5-Math-7B. Selected by maximum consistency with $(\lambda_\text{SFT}, \lambda_\text{RLVR})=(0.79,1.82)$.}
        \label{fig:dyst-Qwen2.5-Math-7B}
    \end{subfigure}
    \hfill
    \begin{subfigure}[t]{0.49\textwidth}
        \centering
        \includegraphics[width=\linewidth]{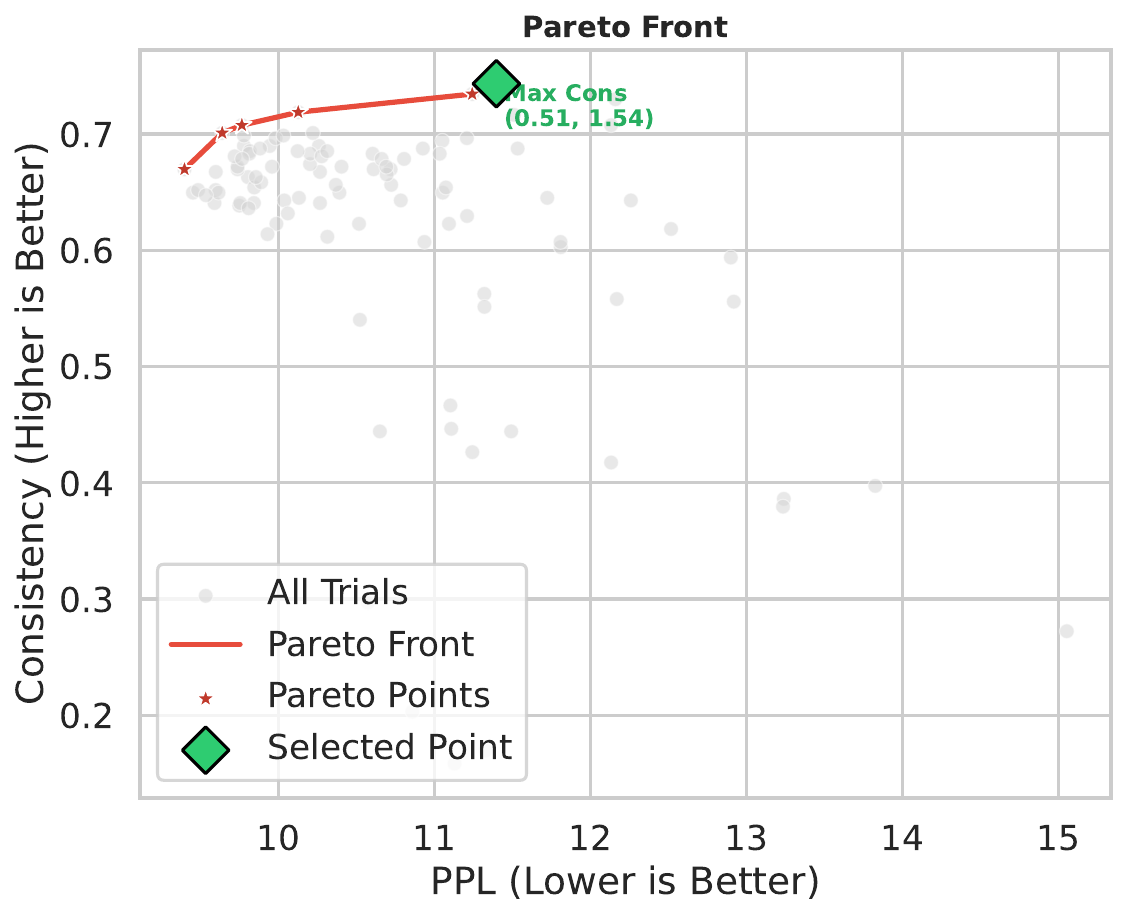}
        \caption{\small Qwen2.5-Math-1.5B. Selected by maximum consistency with $(\lambda_\text{SFT}, \lambda_\text{RLVR})=(0.51,1.54)$.}
        \label{fig:dyst-Qwen2.5-Math-1.5B}
    \end{subfigure}
    \hfill
    \begin{subfigure}[t]{0.49\textwidth}
        \centering
        \raisebox{0.3em}
        {\includegraphics[width=\linewidth]{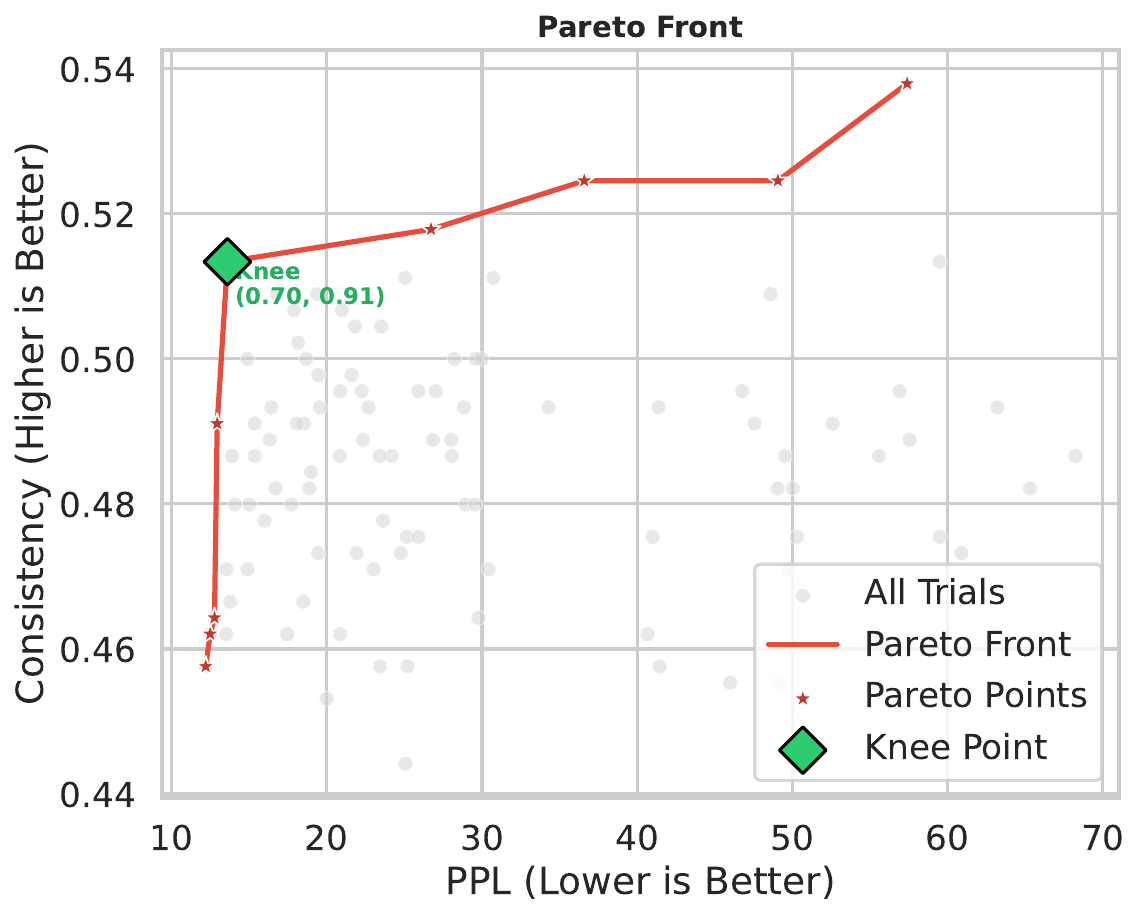}}
        \caption{\small LLaMA3.1-8B. Selected by the knee point with $(\lambda_\text{SFT}, \lambda_\text{RLVR})=(0.70,0.91)$.}
        \label{fig:dyst-filtered-LLaMA-3.1-8B}
    \end{subfigure}
    \hfill
    \begin{subfigure}[t]{0.49\textwidth}
        \centering
        \includegraphics[width=\linewidth]{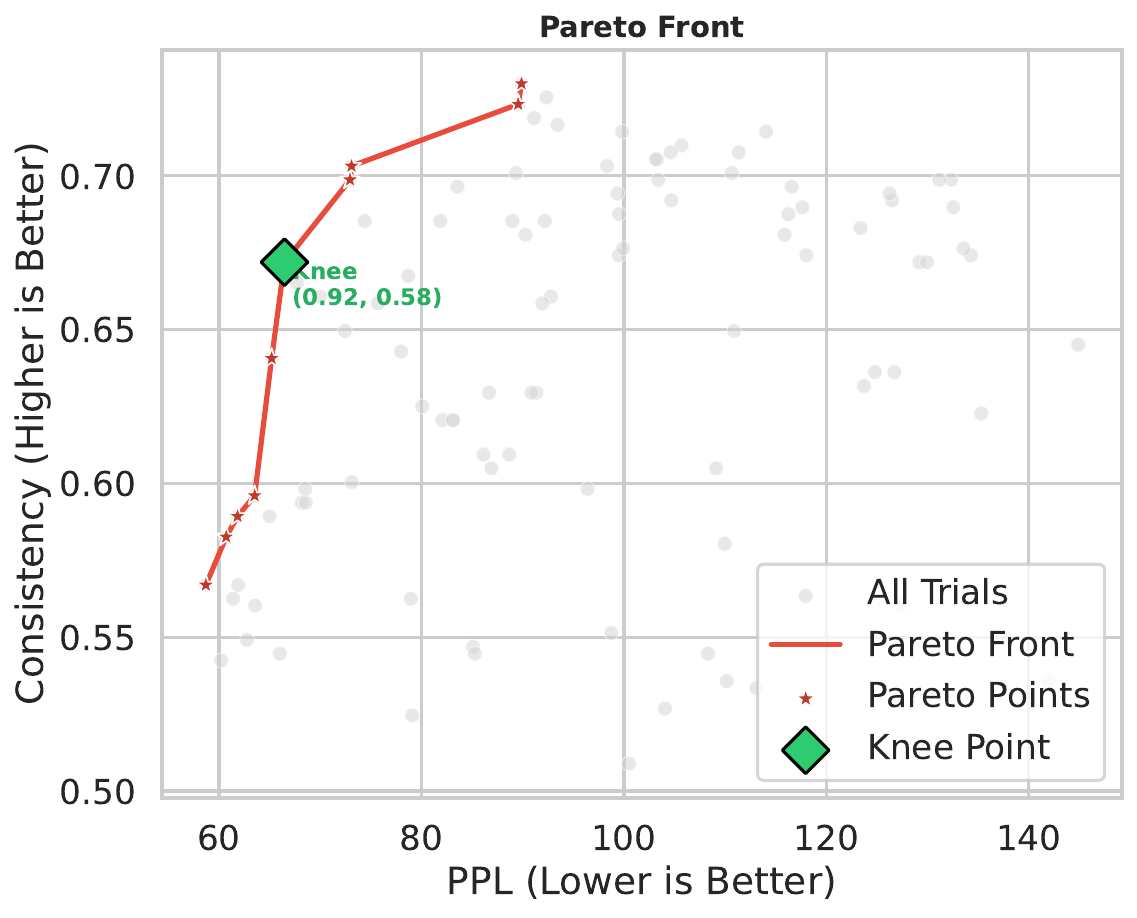}
        \caption{\small LLaMA3.1-8B with SFT++. Selected by the knee point with $(\lambda_\text{SFT}, \lambda_\text{RLVR})=(0.92,0.58)$.}
        \label{fig:dyst-no-filtered-LLaMA-3.1-8B}
    \end{subfigure}
    \caption{\small Pareto frontiers learned by \ours across different backbones. Each point corresponds to one candidate coefficient pair. We select the final coefficients by maximum consistency for Qwen-series models and by the knee point for LLaMA3.1-8B.
    }
    \vspace{-3ex}
    \label{fig:pareto_frontiers}
\end{figure*}

\section{Experimental Details}
\label{appendix:experimental_details}

\subsection{Coefficient Optimization Configuration for \ours}

To optimize the merging coefficients $\lambda_\text{SFT}$ and $\lambda_\text{RLVR}$, we use Optuna~\cite{akiba2019optuna} with the Tree-structured Parzen Estimator (TPE) sampler. We run 100 trials for each setting. Since each trial requires generation from a candidate merged model, we use vLLM~\cite{kwon2023efficient} for efficient inference. The generation hyperparameters follow the evaluation setting: temperature 0.6, maximum generation length 8192 for Qwen-series models, and maximum generation length 2048 for LLaMA3.1-8B. The search space for both coefficients is $[0,2]$, which we find sufficient to identify stable and effective combinations.

\subsection{Model Selection from the Pareto Frontier}
\label{appendix:pareto_front}

Fig.~\ref{fig:pareto_frontiers} visualizes the Pareto frontiers learned by \ours across different backbones.
For stronger backbones, such as the Qwen-series models, the Pareto front usually lies within a narrow perplexity range.
We therefore select the point with the highest consistency.
For weaker backbones, such as LLaMA3.1-8B, high consistency can sometimes coincide with degraded perplexity.
In this case, we select the knee point on the Pareto frontier, computed by minimizing the normalized Euclidean distance to the ideal point with maximum consistency and minimum perplexity.

\subsection{Implementation of Source Models}

For Qwen2.5-Math-7B, we use the public checkpoints released by LUFFY\footnote{\url{https://huggingface.co/collections/Elliott/luffy-rl}}.
For Qwen2.5-Math-1.5B under the DFT configuration~\cite{wu2025generalization}, we use the public DFT checkpoints\footnote{\url{https://huggingface.co/collections/Liang0223/dft}}.
For the remaining settings, including Qwen2.5-Math-1.5B under the LUFFY configuration and LLaMA3.1-8B, the corresponding source weights are not publicly released.
We therefore train the SFT and RLVR source models ourselves.

\paragraph{RLVR implementation.}
Our RLVR implementation is based on VeRL\footnote{\url{https://github.com/verl-project/verl}}.
Following LUFFY~\cite{yan2025learning}, we use GRPO without the KL divergence term. We set the rollout batch size to 128, update batch size to 64, number of rollouts to 8, learning rate to $1\times10^{-6}$, and rollout temperature to 1.0. Training runs for 500 steps on one node with $8\times$ A800-80GB GPUs.

\paragraph{SFT implementation.}
We use LLaMA-Factory~\cite{zheng2024llamafactory} for full fine-tuning. The global batch size is 64. We use learning rate $5\times10^{-5}$, cosine scheduler with warmup ratio 0.1, DeepSpeed ZeRO-3, bfloat16 precision, and 3 training epochs.

\subsection{System prompts}
For Qwen-series models, we use the same system prompt for both training and inference. The prompt asks the model to produce a structured reasoning process followed by a final answer.
\begin{tcolorbox}[
    center,
    arc=0mm,
    boxrule=1pt,
    colback=blue!6!white,
    colframe=black,
    colbacktitle=black,
    attach boxed title to top left={yshift=-0.1in,xshift=0.15in},
    boxed title style={boxrule=0pt,colframe=white}
]
Your task is to follow a systematic, thorough reasoning process before providing the final solution. This involves analyzing, summarizing, exploring, reassessing, and refining your thought process through multiple iterations. Structure your response into two sections: Thought and Solution. In the Thought section, present your reasoning using the format: “\verb|<think>\n| {thoughts} \verb|</think>\n|”. Each thought should include detailed analysis, brainstorming, verification, and refinement of ideas. After “\verb|</think>\n|” in the Solution section, provide the final, logical, and accurate answer, clearly derived from the exploration in the Thought section. If applicable, include the answer in \verb|\boxed{}| for closed-form results like multiple choices or mathematical solutions. \\
\textbf{User:} This is the problem: \verb|{QUESTION}| \\
\textbf{Assistant:} \verb|<think>|
\end{tcolorbox}

For LLaMA3.1-8B, we use different prompts for SFT and RLVR training. For SFT training, we use the same prompt as the Qwen-series models. For RLVR training, we follow prior work~\cite{yan2025learning}, which finds that LLaMA models are less stable under complex system prompts. We therefore use a simpler Chain-of-Thought prompt that omits the \verb|</think>| delimiter.
\begin{tcolorbox}[
    center,
    arc=0mm,
    boxrule=1pt,
    colback=blue!6!white,
    colframe=black,
    colbacktitle=black,
    attach boxed title to top left={yshift=-0.1in,xshift=0.15in},
    boxed title style={boxrule=0pt,colframe=white}
]
\textbf{User: }\verb|{QUESTION}| \\
\textbf{Answer:} Let's think step by step.
\end{tcolorbox}

During inference with LLaMA3.1-8B, we use a hybrid prompt to better align with both the SFT and RLVR source models. This prompt keeps the structured \verb|<think>| format used by SFT while also including the simpler CoT prefix used during RLVR training.
\begin{tcolorbox}[
    center,
    arc=0mm,
    boxrule=1pt,
    colback=blue!6!white,
    colframe=black,
    colbacktitle=black,
    attach boxed title to top left={yshift=-0.1in,xshift=0.15in},
    boxed title style={boxrule=0pt,colframe=white}
]
Your task is to follow a systematic, thorough reasoning process before providing the final solution. This involves analyzing, summarizing, exploring, reassessing, and refining your thought process through multiple iterations. Structure your response into two sections: Thought and Solution. In the Thought section, present your reasoning using the format: “\verb|<think>\n| {thoughts} \verb|</think>\n|”. Each thought should include detailed analysis, brainstorming, verification, and refinement of ideas. After “\verb|</think>\n|” in the Solution section, provide the final, logical, and accurate answer, clearly derived from the exploration in the Thought section. If applicable, include the answer in \verb|\boxed{}| for closed-form results like multiple choices or mathematical solutions. \\
\textbf{User:} This is the problem: \verb|{QUESTION}| \\
\textbf{Assistant:} \verb|<think>|\\
\textbf{Answer:} Let's think step by step.
\end{tcolorbox}

\subsection{Baseline Implementations}
For SimpleRL-Zero, OpenReasoner-Zero, PRIME-Zero, and Oat-Zero, we report the results from LUFFY~\cite{yan2025learning}. For baselines with public weights, including SFT, On-Policy RL, SFT+RL, and LUFFY, we download the released checkpoints and re-evaluate them under our evaluation setting. For model merging baselines, including TIES-Merging and DARE, we use the same source models as \ours and follow their official hyperparameter settings, with drop rate 10\%.

\section{Algorithm}
Algo.~\ref{alg:dyst} summarizes the full procedure of \ours.

\begin{algorithm}[tb]
   \caption{Procedure of Decoupled Test-time Synthesis (\ours)}
   \label{alg:dyst}
\begin{algorithmic}[1]
   \STATE {\bfseries Input:} Base model $\theta_\text{base}$, SFT model $\theta_\text{SFT}$, RLVR model $\theta_\text{RLVR}$, unlabeled query pool $\mathcal{D}_\text{pool}$.
   \STATE {\bfseries Hyperparameters:} Sparsity ratio $p=30\%$, adaptation set size $n=64$, optimization trials $T=100$.
   \STATE {\bfseries Output:} Final merged model $\theta^*$.
   \STATE
   \STATE \textbf{Stage 1: Difficulty-Aware Data Selection}
   \FOR{each query $x \in \mathcal{D}_\text{pool}$}
       \STATE Compute consistency scores $C_{\theta_\text{SFT}}(x)$ and $C_{\theta_\text{RLVR}}(x)$ using Eq.~\eqref{eq:consistency}.
       \STATE Compute difficulty score $D(x)=1-\frac{1}{2}(C_{\theta_\text{SFT}}(x)+C_{\theta_\text{RLVR}}(x))$.
   \ENDFOR
   \STATE Filter out queries with $D(x)>0.8$.
   \STATE Split the remaining queries into low-difficulty and medium-difficulty pools at the median of $D(x)$.
   \STATE Sample equally from the two pools to form $\mathcal{D}_\text{sel}$ with $n=64$ queries.
   \STATE
   \STATE \textbf{Stage 2: Selective Task-Vector Sparsification}
   \STATE Extract task vectors: $\tau_\text{SFT}\leftarrow \theta_\text{SFT}-\theta_\text{base}$ and $\tau_\text{RLVR}\leftarrow \theta_\text{RLVR}-\theta_\text{base}$.
   \FOR{each vector $\tau \in \{\tau_\text{SFT}, \tau_\text{RLVR}\}$}
       \STATE Create mask $M$ that retains the top-$p\%$ entries by magnitude.
       \STATE Compute rescaling factor $\gamma=\|\tau\|_2/(\|\tau\odot M\|_2+\epsilon)$.
       \STATE Obtain processed vector $\tilde{\tau}\leftarrow \gamma(\tau\odot M)$.
   \ENDFOR
   \STATE
   \STATE \textbf{Stage 3: Bayesian Coefficient Optimization}
   \STATE Initialize search space $\lambda_\text{SFT},\lambda_\text{RLVR}\in[0,2]$.
   \FOR{$t=1$ to $T$}
       \STATE Sample coefficients $\lambda_\text{SFT}^{(t)},\lambda_\text{RLVR}^{(t)}$ using Bayesian optimization.
       \STATE Construct candidate model:
       $\theta^{(t)}=\theta_\text{base}+\lambda_\text{SFT}^{(t)}\tilde{\tau}_\text{SFT}+\lambda_\text{RLVR}^{(t)}\tilde{\tau}_\text{RLVR}$.
       \STATE Evaluate consistency $C^{(t)}$ and perplexity $\text{PPL}^{(t)}$ on $\mathcal{D}_\text{sel}$.
       \STATE Update the Bayesian optimizer with $(C^{(t)},\text{PPL}^{(t)})$.
   \ENDFOR
   \STATE
   \STATE \textbf{Stage 4: Model Selection}
   \STATE Construct Pareto frontier $\mathcal{F}$ from all trials, maximizing consistency and minimizing perplexity.
   \IF{the backbone is strong, such as Qwen2.5-Math}
       \STATE Select $\theta^*$ from $\mathcal{F}$ with the highest consistency.
   \ELSE
       \STATE Select $\theta^*$ from $\mathcal{F}$ at the knee point.
   \ENDIF
   \STATE \textbf{return} $\theta^*$
\end{algorithmic}
\end{algorithm}

\section{Additional Revision Experiments}
\label{appendix:revision_experiments}

This section provides additional analyses for Sec.~\ref{sec:analysis} and extended ablations for Sec.~\ref{sec:ablation}. Unless otherwise stated, all results use Qwen2.5-Math-7B.

\subsection{Robustness of Data Selection}
\label{appendix:data_selection_robustness}

We examine the sensitivity of \ours to two data-selection choices: the threshold for filtering overly hard queries and the easy-to-medium ratio used for stratified sampling.

As shown in Tab.~\ref{tab:difficulty_threshold}, the performance is stable when the threshold is between 0.6 and 0.8. When all hard queries are retained, namely $D(x)\leq1.0$, the average score drops to 47.7. This suggests that queries that are unstable for both source models can introduce noise into coefficient search.

As shown in Tab.~\ref{tab:easy_med_ratio}, Balanced sampling achieves the best average score. Low-difficulty queries provide stable signals, while medium-difficulty queries offer stronger discrimination between coefficient configurations. And as reported in Tab.~\ref{tab:adaptation_selection}, the balanced strategy achieves the best average score and the strongest results on AIME and AMC. This indicates that stratifying by difficulty provides a more useful adaptation set than random sampling or single-difficulty selection.

\begin{table}[t]
\centering
\caption{Difficulty threshold sweep for filtering overly hard adaptation queries.}
\renewcommand{\arraystretch}{0.98}
\resizebox{0.85\linewidth}{!}{%
\begin{tabular}{lccccccc}
\toprule
\textbf{Threshold} & \textbf{AIME 24} & \textbf{AIME 25} & \textbf{AMC} & \textbf{MATH500} & \textbf{Minerva} & \textbf{Olympiad} & \textbf{Average} \\
\midrule
0.2 & 27.8 & 20.0 & 59.8 & 87.2 & 42.3 & 52.4 & 48.3 \\
0.4 & 27.9 & 19.9 & 62.0 & 86.8 & 43.4 & 54.5 & 49.1 \\
0.6 & 29.9 & 19.8 & 61.9 & 88.2 & 43.8 & 52.6 & \textbf{49.3} \\
0.8 & 32.9 & 23.8 & 63.2 & 86.8 & 42.6 & 48.3 & \textbf{49.3} \\
1.0 & 28.6 & 19.2 & 60.1 & 85.0 & 41.2 & 51.9 & 47.7 \\
\bottomrule
\end{tabular}%
}
\label{tab:difficulty_threshold}
\end{table}

\begin{table}[t]
\centering
\caption{Ablation on the easy-to-medium ratio used to construct the adaptation set.}
\renewcommand{\arraystretch}{0.98}
\resizebox{0.85\linewidth}{!}{%
\begin{tabular}{lccccccc}
\toprule
\textbf{Easy:Medium} & \textbf{AIME 24} & \textbf{AIME 25} & \textbf{AMC} & \textbf{MATH500} & \textbf{Minerva} & \textbf{Olympiad} & \textbf{Average} \\
\midrule
0:100 & 30.8 & 19.4 & 59.8 & 87.8 & 43.4 & 51.7 & 48.8 \\
25:75 & 29.9 & 18.5 & 61.7 & 85.6 & 44.9 & 48.9 & 48.3 \\
50:50 & 32.9 & 23.8 & 63.2 & 86.8 & 42.6 & 48.3 & \textbf{49.3} \\
75:25 & 28.8 & 20.2 & 60.5 & 86.2 & 41.2 & 53.9 & 48.5 \\
100:0 & 31.8 & 23.3 & 62.6 & 85.4 & 41.2 & 50.4 & 49.1 \\
\bottomrule
\end{tabular}%
}
\label{tab:easy_med_ratio}
\end{table}

\begin{table}[t]
\centering
\caption{Comparison of alternative adaptation-set selection strategies.}
\renewcommand{\arraystretch}{0.98}
\resizebox{0.92\linewidth}{!}{%
\begin{tabular}{lccccccc}
\toprule
\textbf{Strategy} & \textbf{AIME 24} & \textbf{AIME 25} & \textbf{AMC} & \textbf{MATH500} & \textbf{Minerva} & \textbf{Olympiad} & \textbf{Average} \\
\midrule
Random & 27.4 & 19.9 & 59.9 & 86.2 & 41.9 & 52.7 & 48.0 \\
Diverse (equal per source) & 26.4 & 18.4 & 61.9 & 85.8 & 43.0 & 51.6 & 47.8 \\
Easy Only & 31.8 & 23.3 & 62.6 & 85.4 & 41.2 & 50.4 & 49.1 \\
Medium Only & 30.8 & 19.4 & 59.8 & 87.8 & 43.4 & 51.7 & 48.8 \\
Ours (balanced) & 32.9 & 23.8 & 63.2 & 86.8 & 42.6 & 48.3 & \textbf{49.3} \\
\bottomrule
\end{tabular}%
}
\label{tab:adaptation_selection}
\end{table}

\subsection{Robustness of Sparsification}
\label{appendix:sparsification_robustness}

We next examine the effect of sparsification on standalone task-vector quality and on merging performance under fixed coefficients.

As shown in Tab.~\ref{tab:standalone_sparsity}, aggressive pruning still preserves most standalone performance. For example, retaining only 10--30\% of the entries keeps the performance close to the dense task vectors. This supports the use of sparse task vectors in \ours.

As presented in Tab.~\ref{tab:sparsification_fixed}, under the same coefficients, sparsification improves the average score by 0.9 points. This shows that sparsification contributes beyond coefficient tuning by reducing unnecessary interference before merging.

\begin{table}[t]
\centering
\caption{\small Standalone performance after retaining only the top-$k\%$ parameters in each task vector.}
\renewcommand{\arraystretch}{0.98}
\resizebox{0.25\linewidth}{!}{%
\begin{tabular}{lcc}
\toprule
\textbf{Top-$k\%$} & \textbf{GRPO} & \textbf{SFT} \\
\midrule
10\% & 43.8 & 43.2 \\
20\% & 44.6 & 45.3 \\
30\% & 44.3 & 46.2 \\
40\% & 44.4 & 45.8 \\
50\% & 44.0 & 45.2 \\
60\% & 44.8 & 44.1 \\
70\% & 44.4 & 44.0 \\
80\% & 44.4 & 44.7 \\
90\% & 44.4 & 44.0 \\
100\% & 44.6 & 0.44.2 \\
\bottomrule
\end{tabular}%
}
\label{tab:standalone_sparsity}
\end{table}

\begin{table}[t]
\centering
\caption{Effect of sparsification under the fixed coefficient setting $(1.0,1.0)$.}
\renewcommand{\arraystretch}{0.98}
\resizebox{0.92\linewidth}{!}{%
\begin{tabular}{lccccccc}
\toprule
\textbf{Model} & \textbf{AIME 24} & \textbf{AIME 25} & \textbf{AMC} & \textbf{MATH500} & \textbf{Minerva} & \textbf{Olympiad} & \textbf{Average} \\
\midrule
$(1.0, 1.0)$ w/o Sparsification & 29.4 & 23.9 & 60.7 & 85.4 & 40.4 & 47.3 & 47.8 \\
$(1.0, 1.0)$ w/ Sparsification & 30.4 & 23.9 & 60.8 & 86.0 & 44.9 & 46.5 & \textbf{48.7} \\
\bottomrule
\end{tabular}%
}
\label{tab:sparsification_fixed}
\end{table}

\subsection{Gradient Conflict in Joint Training}
\label{appendix:training_conflict}

To complement the task-vector sign interference in Finding~2, we measure the gradient conflict ratio between SFT and GRPO objectives during joint training. 

The result is reported in Tab.~\ref{tab:grad_conflict}. The conflict ratio remains close to 0.5 throughout training.
This indicates that the directional conflict between SFT and GRPO objectives is persistent under this joint-training setup, rather than only a short-lived optimization artifact.

\begin{table}[t]
\centering
\caption{Gradient conflict ratio between SFT and GRPO, averaged over 10-step windows.}
\renewcommand{\arraystretch}{0.98}
\resizebox{\linewidth}{!}{%
\begin{tabular}{lccccccccccc}
\toprule
\textbf{Steps} & \textbf{1--10} & \textbf{11--20} & \textbf{21--30} & \textbf{31--40} & \textbf{41--50} & \textbf{51--60} & \textbf{61--70} & \textbf{71--80} & \textbf{81--90} & \textbf{91--100} & \textbf{Avg.} \\
\midrule
Mean Conflict Ratio & 0.499 & 0.505 & 0.506 & 0.505 & 0.501 & 0.505 & 0.513 & 0.505 & 0.502 & 0.512 & 0.505 \\
\bottomrule
\end{tabular}%
}\label{tab:grad_conflict}
\end{table}

\subsection{Cross-Architecture Module Analysis}
\label{appendix:cross_architecture}

To test whether the module-wise differences in Finding~3 generalize beyond Qwen2.5-Math-7B, we repeat the activation-distribution analysis on LLaMA3.1-8B. The result is reported in Tab.~\ref{tab:cross_achi_module_analysis}.

Across both backbones, SFT and RLVR show different module-wise activation patterns.
The exact modules with the largest differences vary by architecture, but the broader pattern remains consistent: the two post-training paradigms do not concentrate their largest updates in identical ways.
This supports the view that SFT and RLVR provide partly complementary update signals.

\begin{table}[t]
\centering
\caption{\small Module-wise activated parameter ratios among the top-10\% entries across Qwen2.5-Math-7B and LLaMA3.1-8B.}
\renewcommand{\arraystretch}{0.98}
\resizebox{0.78\linewidth}{!}{%
\begin{tabular}{lcccc}
\toprule
\textbf{Module} & \textbf{Qwen-7B SFT} & \textbf{Qwen-7B RLVR} & \textbf{LLaMA-8B SFT} & \textbf{LLaMA-8B RLVR} \\
\midrule
Attention & 12.30\% & 13.86\% & 10.07\% & 11.82\% \\
Embedding & 0.38\% & 0.59\% & 1.70\% & 1.02\% \\
LM Head & 4.38\% & 0.48\% & 11.69\% & 0.37\% \\
LayerNorm & 19.90\% & 14.61\% & 10.17\% & 0.07\% \\
MLP & 11.72\% & 11.25\% & 10.78\% & 11.83\% \\
\bottomrule
\end{tabular}%
}
\label{tab:cross_achi_module_analysis}
\end{table}

\subsection{Harder Samples on Weaker Backbones}
\label{appendix:weaker_backbones}

For LLaMA3.1-8B, we further test whether training-based integration can directly benefit from harder samples, or whether decoupled synthesis is needed.

From the result presented in Tab.~\ref{tab:harder_sample}, we can observe that directly adding harder samples to LUFFY leads to near-zero performance in this setting. In contrast, \ours can use SFT++ as a stronger knowledge source and synthesize it with the RLVR checkpoint at test time. This supports the advantage of decoupled synthesis on weaker backbones, where direct training-time integration can be unstable.

\begin{table}[t]
\centering
\caption{\small LLaMA3.1-8B performance when harder long-generation samples with length $>2048$ tokens are included.}
\renewcommand{\arraystretch}{0.98}
\resizebox{0.92\linewidth}{!}{%
\begin{tabular}{lccccccc}
\toprule
\textbf{Model} & \textbf{AIME 24} & \textbf{AIME 25} & \textbf{AMC} & \textbf{MATH500} & \textbf{Minerva} & \textbf{Olympiad} & \textbf{Average} \\
\midrule
LUFFY & 1.9 & 0.1 & 13.5 & 39.0 & 15.1 & 9.6 & 13.2 \\
LUFFY$^\star$ & 0.0 & 0.0 & 0.0 & 0.2 & 0.1 & 0.0 & 0.0 \\
\textbf{\ours (SFT++ + On-Policy RL)} & 3.3 & 8.3 & 31.2 & 55.6 & 12.5 & 24.4 & \textbf{22.6} \\
\bottomrule
\end{tabular}%
}
\label{tab:harder_sample}
\end{table}


\end{document}